\documentclass[a4paper,fleqn]{cas-dc}

\usepackage{bbm}
\usepackage[font=footnotesize,labelfont=rm,textfont=rm]{subfig}
\usepackage[authoryear]{natbib}
\usepackage[normalem]{ulem}
\useunder{\uline}{\ul}{}

\usepackage{pifont}
\usepackage{CJK}
\def\tsc#1{\csdef{#1}{\textsc{\lowercase{#1}}\xspace}}
\tsc{WGM}
\tsc{QE}
\tsc{EP}
\tsc{PMS}
\tsc{BEC}
\tsc{DE}

\usepackage{color, xcolor, colortbl}
\definecolor{Gray}{gray}{0.9}

\begin{document}

\let\WriteBookmarks\relax
\def\floatpagepagefraction{1}
\def\textpagefraction{.001}
\shorttitle{SSPFusion: A semantic structure-preserving approach for multi-modality image fusion}
\shortauthors{Q. Yang et~al.}

\title [mode = title]{SSPFusion: A semantic structure-preserving approach for multi-modality image fusion}                      

\author[1,4]{Qiao Yang}[orcid=0009-0007-8234-9239]
\ead{22140575@bjtu.edu.cn}

\author[2]{Yu Zhang}[orcid=0000-0001-5728-7645]
\ead{uzeful@buaa.edu.cn}

\author[3]{Yutong Chen}[orcid=0009-0005-6142-8822]
\ead{yutochen30@gmail.com}

\author[1]{Jian Zhang}[orcid=0000-0002-9804-951X]
\cormark[1]
\ead{jianzh@bjtu.edu.cn}

\author[1]{Shunli Zhang}[orcid=0000-0002-8186-8949]
\ead{slzhang@bjtu.edu.cn}

\affiliation[1]{organization={ School of Software Engineering, Beijing Jiaotong University},
    addressline={Beijing 100044, China}}
\affiliation[2]{
organization={School of Astronautics, Beihang University},
    addressline={Beijing 100083, China}}
\affiliation[3]{
organization={Faculty of Engineering and Information Technology, University of Melbourne},
    addressline={Parkville 3010, Australia}}
\affiliation[4]{
organization={Ordos Laboratory, Ordos},
    addressline={Inner Mongolia, 017010, China}}

%\fntext[\cor]{Corresponding authors: Malu Zhang, Liang-Jian Deng}
\cortext[1]{Corresponding author: Jian Zhang; Co-first author: Yu Zhang}

% \address[1]{School of Computer Science and Technology, China University of Mining and Technology, Xuzhou 221116, China}
% \address[2]{Mine Digitization Engineering Research Center of the Ministry of Education, Xuzhou 221116, China}
% \address[3]{Innovation Research Center of Disaster Intelligent Prevention and Emergency Rescue, Xuzhou 221116, China}
% \address[4]{School of Electrical Engineering and Computer Science, University of Ottawa, Ottawa, ON K1N 6N5, Canada}

% \cortext[cor1]{Corresponding author}

\begin{abstract}
Most existing learning-based  multi-modality image fusion (MMIF) methods suffer from significant structure inconsistency due to their inappropriate usage of structural features at the semantic level.
To alleviate these issues, we propose a semantic structure-preserving fusion approach for MMIF, namely SSPFusion. 
At first, we design a structural feature extractor (SFE) to extract the prominent structural features from multiple input images. 
Concurrently, we introduce a transformation function with Sobel operator to generate self-supervised structural signals in these extracted features.
Subsequently, we design a multi-scale structure-preserving fusion (SPF) module, guided by the generated structural signals, to merge the structural features of input images. This process ensures the preservation of semantic structure consistency between the resultant fusion image and the input images. Through the synergy of these two robust modules of SFE and SPF, our method can generate high-quality fusion images and demonstrate good generalization ability.
Experimental results, on both infrared-visible image fusion and medical image fusion tasks, demonstrate that our method outperforms nine state-of-the-art methods in terms of both qualitative and quantitative evaluations. Code of this work will be released at~\href{https://github.com/QiaoYang-CV/SSPFUSION}{https://github.com/QiaoYang-CV/SSPFUSION}.
\end{abstract}

\begin{keywords}
Image Fusion \sep Infrared and Visible Image \sep Medical Image \sep Self-Supervised Learning \sep Structure Consistency
\end{keywords}

\maketitle

\section{Introduction}\label{sec1}
Multi-Modality Image Fusion (MMIF) aims to integrate the underlying features of input images from multiple modalities into one comprehensive image, making it a pivotal subject within computer vision.
MMIF encompasses several specialized subfields. Infrared-Visible Image Fusion (IVIF) ~\citep{ZHANG2017227, Liu-reviwer-1} aims to preserve both thermal radiation information in the input infrared images and detailed texture information in the input visible images. The fused images effectively mitigate the limitations associated with visible images, such as their sensitivity to illumination conditions, as well as the inherent noisiness and low-resolution characteristic of infrared images. 
Meanwhile, Medical Image Fusion (MIF) ~\citep{mif1} aims to clearly display various abnormalities by integrating multiple medical imaging modalities.
Remote Sensing Image Fusion (RSIF)~\citep{RSIF} aims to combine complementary information from panchromatic (PAN) images, which provide high spatial resolution, with hyperspectral (HSI) or multispectral (MSI) images, which offer rich spectral signatures. Additionally, it may integrate data from other modalities such as synthetic aperture radar (SAR) for all-weather penetration capability or LiDAR for precise elevation modeling. It can be concluded that MMIF tasks inherently preserve a multi-modality nature, as the infrared-visible pairs in IVIF, the multi-modality medical scans (e.g. CT-MRI, PET-MRI and SPECT-MRI) in MIF, and the multi-spectral satellite imagery (e.g. PAN-HSI) in RSIF. Furthermore, the fused image can provide a foundation for various downstream computer-vision tasks, e.g., multi-target detection~\citep{TarDAL}, real-time semantic egmentation~\citep{Rethinkinghighlevel}, and medical diagnosis~\citep{mif1, eswa_2_Medical}, etc. 
Although RSIF is equally important, its geospatial-specific challenges often necessitate standalone treatment. Therefore, this paper focuses on IVIF and MIF for in-depth analysis.

\begin{figure}[!t]
    \centering
    \includegraphics[width=1.0 \columnwidth]{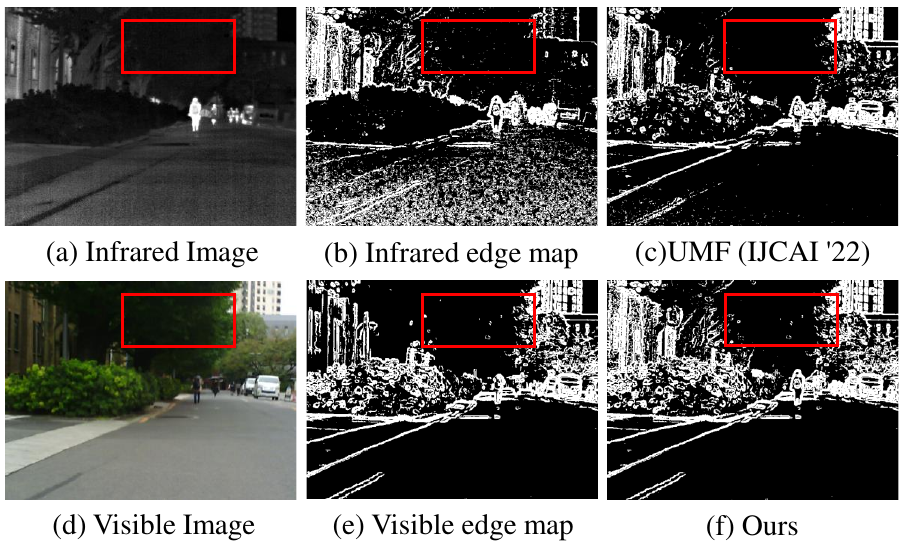}
    \caption{A pair IVIF image of (a) and (d), from MSRS, fused by UMF~\citep{UMF-CMGR} method (c) and our method (f). As shown in the red box, our method can obtain robust cross-modal representation from the edge maps (b) and (e).}
    \label{fig:1}
\end{figure}

Recently, many learning-based MMIF methods have been developed~\citep{U2Fusion, DeFusion, ESWA_1_ivif}. The major barrier to producing high-quality fused images is the structural inconsistency, so that a common strategy~\citep{fusiongan, U2Fusion} is to maximize the structural similarity between the fused and input images.
However, as shown in the red boxes of Figure~\ref{fig:1}, despite the tree being clearly shown in the visible image, it is absent in the fused image of the state-of-the-art learning-basedd IVIF method UMF~\citep{UMF-CMGR}.
Therefore, some works ~\citep{Edge_4} have noticed the effect of employing structural information, e.g., edge, to promote the fusion. Edge guides the fusion by distinguishing between different parts in the multiple input regions. Unfortunately, the corresponding structure modeling results of those works are not satisfying due to the incompleteness in the edge areas caused by severely poor detail loss.
Similar challenges also persist in the MIF task~\citep{EMFusion}, reflecting an overuse of structural information where the edge map derived from input images implicitly guides the fusion process. Consequently, the above two approaches might have a reduced impact on preserving higher-level semantic structures.

\begin{figure}[!t]
    \centering
    \includegraphics[width=1.0 \columnwidth]{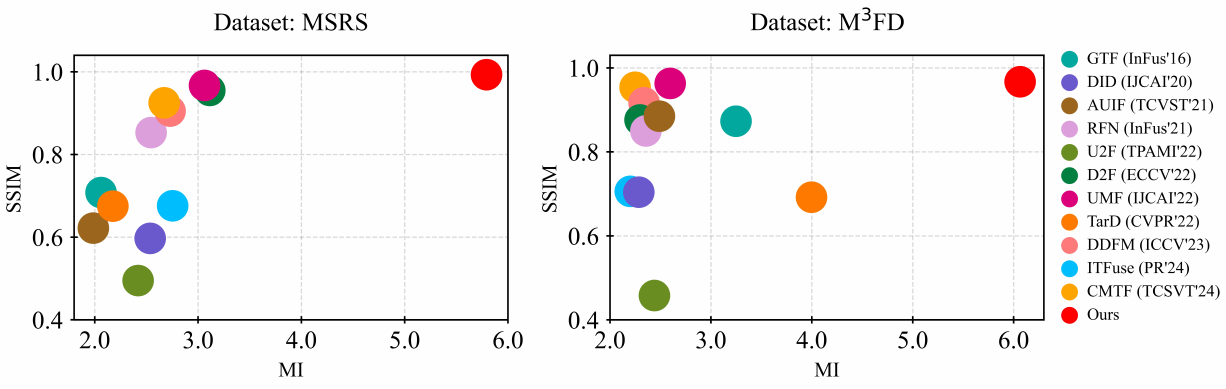}
    \caption{Our method consistently achieves stat-of-the-art performance on datasets: MSRS, M$3$FD.}
    \label{fig: scatter}
\end{figure}

To address the above issues, we formulate a novel multi-modality image fusion paradigm, which introduces self-supervised learning method to allow the image fusion model to focus on the salient features and semantic information rather than low-level edges, so that better structural consistency can be maintained.
Specifically, we propose a semantic structure-preserving approach for MMIF, namely SSPFusion, to explicitly preserve structural information from different modalities. 
we emphasize the explicit modeling of structural details in MMIF. First, we designed a Structural Feature Extractor (SFE) to extract deep structural features, which focus more low-frequency structures and high-frequency semantic information compared to existing methods that rely solely on edge maps. Moreover, we designed a novel Structure-Preserving Fusion (SPF) module to leverage these structural features during the fusion process by using the Sobel operator on the self-supervised structural signal.
Finally, owing to the mutual coordination of the SFE and SPF modules, the proposed method can effectively transfer a considerable amount of information and ultimately preserve the structural features from input images to the fused image.
Experimental results show that our method achieves state-of-the-art performance on both MI and SSIM metrics with the same architecture on multiple datasets, as shown in Figure~\ref{fig: scatter}. In summary, the contributions of this work are mainly threefold:
\begin{itemize} 
    \item We propose a novel self-supervised MMIF framework, which adaptively utilizes multi-modality structura features to efficiently produce fusion images with relieved information loss and superior structural fidelity.
    \item We propose the SFE module for extracting the salient structural features and the SPF module for integrating multi-scale structural features to ensure consistent semantic structure across multiple modalities. 
    \item Extensive experiments are conducted on four datasets for both IVIF and MIF, showing the effectiveness and good generalization ability of our framework.
\end{itemize}

The remainder of this paper is organized as follows: The related work about IVIF are given in Section~\ref{sec:2}. Section~\ref{sec:3} provides a detailed introduction to proposed SSPFusion framework, including Structural Feature Extractor (SFE) and Structure-Preserving Fusion (SPF) module. Section~\ref{sec:4} presents extensive experimental results on three datasets. Finally, Section~\ref{sec:5} concludes this paper.

\section{Related Works}\label{sec:2}
In MMIF tasks, using priors and features from different modalities is a crucial component of efficient learning-based training procedures. Therefore, many techniques have been developed for learning joint representations of multiple input images. In this section, we provide a comprehensive overview of traditional methods and deep learning-based methods in MMIF.

\subsection{Traditional MMIF}
Over the past two decades, numerous traditional fusion algorithms have been proposed to advance  MMIF techniques. These algorithms can generally be categorized into six groups: multi-scale transform-based methods~\citep{2024Infrared}, subspace-based methods~\citep{sub-space1}, saliency detection-based methods~\citep{saliency_detection1}, sparse representation-based methods~\citep{sparse_representation2, sp3, ESWA_1_medical}, optimization-based methods~\citep{GTF, possion}, and hybrid methods~\citep{liu2015general}. Among them, traditional optimization-based learning methods, exemplified by Gradient Transfer Fusion (GTF)~\citep{GTF}, have achieved commendable integration outcomes. 
Despite their strong performance in certain scenarios, existing methods for image fusion encounter several critical limitations: (1) The success of these algorithms heavily relies on hand-crafted feature extractors. The constrained capability of these extractors to capture relevant features hinders the overall effectiveness of the fusion process. (2) When faced with challenging conditions such as cloud cover, fog, precipitation, dim lighting, or overexposure, these methods' performance 
degrades markedly. The resulting fused images are prone to display issues like blurring, haloing effects, and a lack of fine details, which collectively lead to a less than optimal visual outcome.

\subsection{Deep learning methods of MMIF} 

With the advent and progression of deep learning~\citep{chen1, chen2}, more superior fusion models have been developed. 
The current deep learning-based MMIF methods mainly consist of three categories, i.e., pre-trained model-based methods, end-to-end based methods and multi-modality large language model~(MMLLM)-based methods. In this section, we briefly overview those deep learning-based methods.

\textbf{Pre-trained model-based method}. The pre-trained model method aims to establish intrinsic connections among various image tasks~\citep{per2}. Specifically, the commonality of these methods lies in the utilization of pre-trained models for the precise extraction of salient features from source images, manual design of pertinent fusion strategies, and subsequent reconstruction of fused images using decoders. Such as ~\cite{per3} designed a feature extractor and fusion framework based on DenseNet-201.~\cite{DensFuse} proposed DenseFuse, which uses MS-COCO datasets to pre-train the encoder and decoder and two different fusion strategies (l1-norm and addition) to perform feature fusion.
However, pre-trained network models are primarily designed for visual tasks such as image classification or image segmentation, which may not be adaptable to the fusion task. 
Moreover, manual intervention is still required in the fusion process. 
The design of fusion strategies needs to be tailored for different feature extraction networks, increasing the difficulty of the fusion algorithm. 
Finally, due to the phenomenon of model degradation in deep neural networks, manual fusion strategies can result in structure feature loss and addition of redundant features, leading to uneven textures in the final fused image.

\textbf{End-to-end model-based method}. In recent years, to consider the overall fusion framework and avoid the manual design of fusion strategies, researchers have proposed a series of end-to-end fusion algorithms. Specifically, these MMIF methods mainly consist of five categories, i.e., auto-encoder (AE)-based methods~\citep{EMFusion, RFN-Nest, joint_1}, convolutional neural network (CNN)-based methods~\citep{DeFusion, U2Fusion, IFCNN, joint_2, ESWA_2_ivif}, 
generative model(GM)-based methods~\citep{FreqGAN, Cross-Scale_tcvst, Reviewer2-1, VDMUFusion}, transformer-based methods~\citep{CDDFuse, ITFuse, CMTFusion}, and task-driven methods~\citep{MulFS-CAP, IAIFNet, mif2}.

\textit{Methods Based on AE}.
The AE-based fusion methods usually focus on designing feature extraction and fusion architectures to obtain superior cross-modal feature representations~\citep{TANG-2}. For example,~\cite{RFN-Nest} designed the fusion strategy of nest connection based on residual learning. To mitigate essential information loss,~\cite{SFA-Fuse} utilizes an encoder and two decoders. These decoders are specifically employed to reconstruct the visible image and the infrared image. Further,~\cite{Xu_AE} adopted a similar approach to train the autoencoder and simultaneously trained a classifier to determine the pixel-level contributions of the source image. They fused features in a pixel-weighted manner, breaking through the bottleneck of deep fusion relying on manual strategies.

\textit{Methods Based on CNN}.
This kind of method has evolved into various types through the design of sophisticated network structures. For instance, to achieve effective global feature representation, researchers~\citep{cnn_2, EAT, Liu-1-1} employed multi-scale feature extraction using convolution kernels of different sizes. Smaller convolution kernels effectively extract low-frequency information, while larger kernels effectively capture high-frequency information, providing an effective representation of global features. Inspired by image decoupling methods,~\cite{IFSepR} utilized convolutions to decompose source images into background and detail layers, performing feature fusion on different layers to obtain shared and unique features, thereby enhancing the information representation of the fused image at various levels. In recent years, with the effective representation of intrinsic differences in image data through contrastive learning, new insights have been injected into the field of image fusion.~\cite{Liu-1-1} constructed positive and negative sample pairs of infrared and visible light images to enhance the saliency of foreground objects and the texture of background objects. Under the constraints of a designed contrastive loss function, they achieved effective separation between different object features and reduced redundant information in the fused features. Additionally, following the success of Neural Architecture Search (NAS) in low-vision tasks, researchers have applied it to the fusion domain to automatically optimize different convolutional structures for the best fusion model~\citep{smoa}. These CNN-based network structures, meticulously designed, have all improved fusion performance.

\textit{Methods Based on GM}.
The inherent lack of ground truth fusion results in MMIF tasks naturally favors the application of unsupervised learning methods. The unsupervised generative models for obtaining fused images are mainly divided into two types: Generative Adversarial Networks (GAN) and Diffusion Models (DM). In 2019, ~\cite{fusiongan} proposed FusionGAN, which was the first to introduce GAN into the field of infrared and visible light image fusion, forcing the generator to focus on visible light texture information through the discriminator. Subsequently,~\cite{GANMcC} addressed the issue that FusionGAN did not balance the information from infrared and visible images well and proposed GANMcC based on a multi-class discriminator to maintain a good balance between the two. However, methods based on GAN have issues such as instability, difficulty in convergence, and lack of interpretability during the training fusion process. To address this, researchers have developed fusion methods based on DMs, using likelihood estimation to implicitly fine-tune the fusion process and alleviate these problems. For example,~\citep{DDFM} proposed DDFM based on diffusion models and the expectation-maximization algorithm, modeling the fusion task as a posterior sampling model and obtaining stable and controllable fused images through likelihood correction.~\cite{Dif_Fusion} aimed to obtain high-fidelity color information by conducting diffusion modeling in the chrominance dimension, achieving high-fidelity reconstruction of the original image with the participation of multi-channel information.

\textit{Methods Based on transformer}.
Transformer is a model architecture for natural language processing, which is also widely used in the field of image fusion due to its excellent ability to model long-range dependencies~\citep{Tang-1, Tang-3}.~\cite{CGTF} combined CNN and Transformer to complement the strengths of different frameworks; CNN is used to focus on local information, while Transformer is used to perceive global information. Similarly,~\cite{CDDFuse} also adopted the same approach for extracting low and high-frequency features in image decomposition and introduced reversible neural blocks to prevent the loss of important information during long-range transmission.~\cite{SwinFusion} incorporated Swin Transformer into the image fusion framework to extract cross-domain structures in features, enabling global modal information interaction.

\textit{Methods Based on task-driven}.~\cite{joint_1} were among the earliest to consider the application of fusion tasks in downstream visual tasks. To retain more high-frequency semantic information and make the fused images suitable for visual tasks such as image registration~\citep{2024}, segmentation and object detection~\citep{SuperFusion, Liu-3-1, Tang-0} proposed SuperFusion, which takes scene segmentation into account during the image fusion process.~\cite{joint_2} cascaded the fusion network with the object detection network and proposed a corresponding image fusion method to improve the performance of the fusion results for object detection tasks.~\cite{TarDAL} formalized image fusion and object detection with a two-tier optimization modeling, proposed a joint training strategy to train the fusion model and detection model, and set up object discriminators and detail discriminators for objects and details, respectively, to jointly learn the complementarity of the source images. Although it introduces a dual discriminator to highlight distinct structures of targets and textural structure of background, somewhat mitigating the problem of structural loss, the improvement effect is negligible.

\textbf{Multi-modality large language model-based method.} With the development of various Vision-Language models, MMLLM-based methods have demonstrated promising applications in the field of image fusion~\citep{Liu-2-1}. These methods utilize natural language descriptions of different image modalities to guide LLM, with the objective of generating fused images that provide comprehensive scene understanding.~\cite{Zhao_2024_ICML} improved fusion performance by incorporating text derived from ChatGPT to guide the fusion algorithm in understanding text-level semantic information.~\cite{Text-IF} employed the CLIP model to provide relevant text semantic features, assisting the fusion network in achieving results tailored to specific tasks or degradation types.~\cite{Instruction-Driven} integrated the LLaMA model to provide task-oriented adaptive regulation instructions, enabling the algorithm to enhance the accuracy of downstream tasks through implicit fusion. However, the fusion process of MMLLM-based methods is often constrained by hardware resource limitations.

\subsection{Comparison with existing approaches}
In summary, most of the methods mentioned above were developed to maximize the structural similarity between the fused and input images by implicitly utilizing a fusion network. 
Regardless of the quality of the obtained fused images, the disregard of explicit modeling of structural details in original areas, which translates to an ineffective acquisition of cross-modal representations, leads to consequently lower MI~\citep{MI} and SSIM~\citep{SSIM} for the fusion results.

In contrast to other methods, our SSPFusion framework introduces a self-supervised learning strategy to guide the fusion process toward semantically meaningful and structurally consistent outputs. Technically, the novelty lies in that our SSPFusion creatively integrates MMIF with self-supervised learning method, meanwhile, it also combines semantic features with cross-modality structural consistency, which enable the proposed framework to achieve better performance than image fusion methods that employ edges as structural information. Extensive results on IVIF and MIF datasets demonstrate that SSPFusion achieves superior performance in both MI and SSIM metrics, validating its effectiveness and generalization capability.

\section{PROPOSED METHOD}\label{sec:3}

\begin{figure*}[]
\centering
\centerline{\includegraphics[width=2.0 \columnwidth]{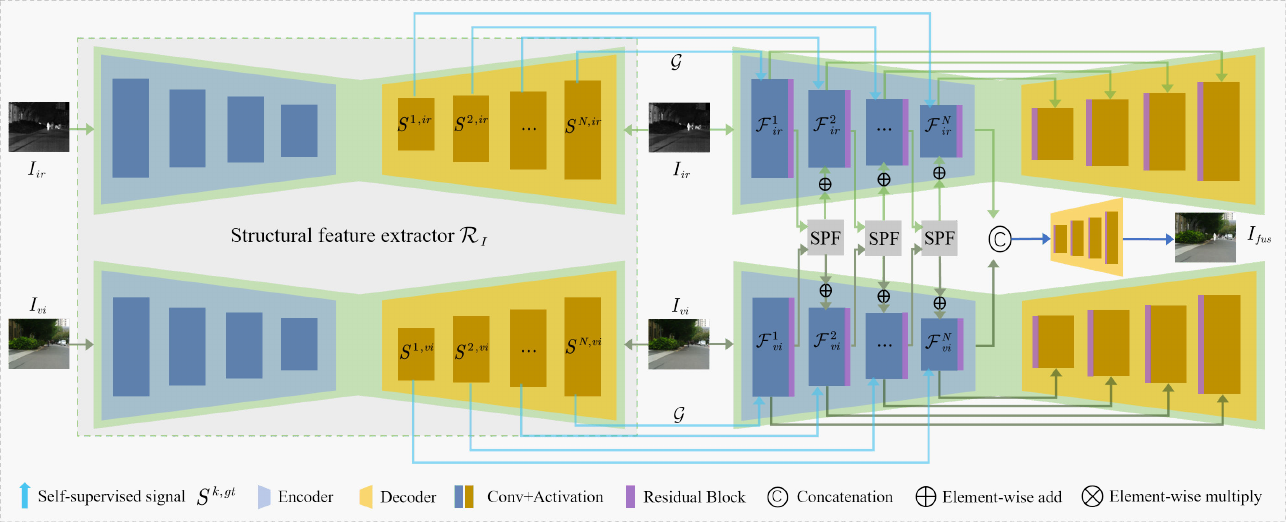}}
\caption{The overall architecture of our proposed multi-modality image fusion method.}
\label{fig:network}
\end{figure*}

As shown in Figure~\ref{fig:network}, the base architecture of our method is a common U-shape~\citep{U-Net} model. 
Firstly, our SFE extracts structure maps as self-supervised signals. Then, guided by these signals, SPF fuses the complementary structures of multiple input images by a structure-preserving mechanism. Finally, the fused image is reconstructed from the fused structural features by the Decoder. For simplicity, we describe SSPFusion using IVIF as an example. Specifically, we denote the infrared, visible, and fused images by $I_{ir}, I_{vi}$ and $I_{fus}$, respectively.
As the YUV color space can effectively separate intensity (representing structure) and color features, all images are transformed into the YUV color space before being fed into the model.

\begin{figure}[]
\centering
\centerline{\includegraphics[width=0.8 \columnwidth]{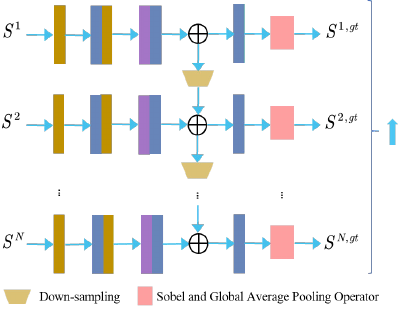}}
\caption{The architecture of structural feature extractor.}
\label{fig:SFE}
\end{figure}

\begin{figure}[]
    \centering
    \includegraphics[width=0.98 \columnwidth]{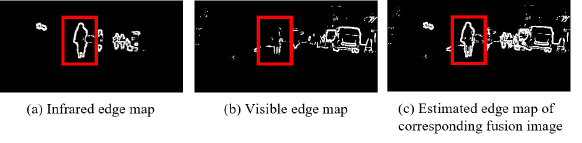}
    \caption{Priori of structure-preserving fusion.}
    \label{fig:prior}
\end{figure}

\begin{figure}[]
\centering
\centerline{\includegraphics[width=0.75 \columnwidth]{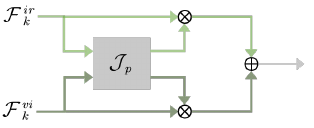}}
\caption{The architecture of structure-preserving fusion.}
\label{fig:JJJ}
\end{figure}

\subsection{Structural feature extractor}
\textbf{Feature extractor}. To effectively extract structural features from the infrared and visible images, a structural feature extractor (SFE) is designed in particular, as illustrated in Figure~\ref{fig:network}. In SFE, the feature maps {$\mathcal{F}_1, ..., \mathcal{F}_N$} are extracted by the multi-layer encoder with a down-sampling operator in each layer, where $N$ is the number of layers in the U-shape model. Specifically, for each layer $k\in[1, N]$, $\mathcal{F}_k$ of the input image $I$ can be extracted by
\begin{equation} 
\setlength{\abovedisplayskip}{6pt}
\mathcal{F}_k=\mathcal{R}_{I}\left( I \right) ,
\setlength{\belowdisplayskip}{6pt}
\end{equation}
where $\mathcal{R}_{I}$ denotes the SFE for recovering structures of $I$. These multi-scale feature maps emphasize the edge regions, benefiting the representation of structural features in the fusion process. 

\textbf{Self-supervision of structures}. 
The large cross-modality variation between multiple input images makes it impractical to straightly bridge their inconsistency in structural feature space. Therefore, we adopt dice loss~\citep{Dice_loss1} to optimize our model supervised by an estimated structure maps of $S^{k,gt}$ to maintain the consistency of multi-modality structural features during fusing images.
Specifically, as illustrated in Figure~\ref{fig:SFE}, we introduce self-supervised signal $S^{k,gt}=\mathcal{G}\left( \mathcal{R}_{I}\left( I\right) \right)$ for SFE (without skip connection) to predict multi-scale structure map $S^{k}$, where $\mathcal{G}$ denotes a transformation function in structure domain, which can be computed by
\begin{equation} 
\setlength{\abovedisplayskip}{6pt}
S_{i,j}^{k,gt}=\left\{ \begin{array}{l}
	1,\ \nabla \mathcal{F}_{k}^{i,j}-\mathcal{A}\mathcal{F}^{i}_{k}<=0\\
	0,\ else\\
\end{array}\right.,
\setlength{\belowdisplayskip}{6pt}
\end{equation}
where $\mathcal{F}^{i,j}_{k}$ is the $j$-th pixel of the $i$-th feature map $\mathcal{F}^{i}_{k}$, $\nabla$ and $\mathcal{A}$ represent the sobel operator and global average pooling operator, respectively. Then, the training loss of structural information is calculated as:
\begin{equation} 
\setlength{\abovedisplayskip}{6pt}
\mathcal{L}_{str}=\sum\limits_{k=1}^N{Dist\left( S^k,S^{k,gt} \right)} ,
\setlength{\belowdisplayskip}{6pt}
\end{equation}
$Dist$ is the dice loss and is defined as 
\begin{equation} 
\setlength{\abovedisplayskip}{6pt}
Dist\left( P,G\right) =\left( \sum\limits_{c}^{H\times W}{p_{c}^{2}+}\sum\limits_c^{H\times W}{g_{c}^{2}} \right) /2\sum\limits_c^{H\times W} ({p_c\times g_c}) ,
\setlength{\belowdisplayskip}{6pt}
\end{equation}
where $p_c$ and $g_c$ are the value of the $c$-th pixel on the predicted structure map $P$ and that of the ground truth $G$, respectively. Note that the dice loss measure is insensitive to foreground/background information, which means it provides structure maps that can alleviate the inconsistency problem.

We use the ground-truth structure maps of the input images as self-supervised signals to supervise the fusion process, paying greater attention to the fusion of semantics structure-preserver while also avoiding the loss of information.

\subsection{Structure-preserving fusion} 
\textbf{Prior knowledge.} As shown in the red boxes of Figure~\ref{fig:prior}, the edge maps (generated by binarizing the images after applying the sobel filter) of infrared and visible images can reflect the object's structures from different views. 
Further, we observe structural information by transforming the input image with the formula
\begin{equation} 
\setlength{\abovedisplayskip}{6pt}
\mathcal{J}_p\left( I_{ir},I_{vi} \right) =\left( 1-\nabla I_{ir} \right) \nabla I_{vi}+\left( 1-\nabla I_{vi} \right) \nabla I_{ir}.
\setlength{\belowdisplayskip}{6pt}
\end{equation}
The output generated by $\mathcal{J}_p$ can retain structures that align with the structure maps derived from the input images.
This inspires us to consider further enhancing the structural features throughout the fusion process.

\textbf{Fusion scheme.} Intuitively, an image integrating the complementary cross-modal structures of infrared and visible images will be helpful for boosting high-level vision tasks.
To this end, we design an effective cross-modal structure-preserving fusion scheme.
Specifically, we first extract the unique structures of $S^{k,ir}$ and $S^{k,vi}$ by: 
\begin{equation}
\setlength{\abovedisplayskip}{6pt}
\mathcal{M}^{k}= \mathcal{J}_p\left (S^{k,ir}, S^{k,vi} \right),
\setlength{\belowdisplayskip}{6pt}
\label{eq:j}
\end{equation}
$\mathcal{M}^{k}$ will be 1 when only one of $S^{k,ir}$ and $S^{k,vi}$ equals to 1. Thus, Eq.~\eqref{eq:j} yields an edge map specifically delineating the unique structures of $S^{k, ir}$ and $S^{k,vi}$, excluding any shared features between them. In fusion images of the conventional-deep-learning-based methods, these unique structures preserved from input images should be significantly less than those from shared features.

Therefore, regions of feature maps (i.e., $S^{k,ir}$ and $S^{k,vi}$) with unique structures should be enhanced appropriately. Specifically, in this work, we enhance them by:
\begin{align}
\hat{S}^{k, ir}=S^{k, ir}+ \left(1-\mathcal{M} _{ir}^{k}\right)\mathcal{F}_{ir}^{k}+\mathcal{M} _{vi}^{k}\mathcal{F}_{ir}^{k},\\
\label{eq:enhance-ir}
\hat{S}^{k, vi}=S^{k, vi}+\mathcal{M} _{ir}^{k}\mathcal{F}_{ir}^{k}+\left(1-\mathcal{M} _{ir}^{k}\right) \mathcal{F}_{ir}^{k}.
\end{align}
\label{eq:enhance-vi}
In this manner, the image model could preserve more complete structural features from the input images into its fused image, as shown in Figure~\ref{fig:JJJ}.
Finally, the fused image $I_{fus}$ with different modalities of structures is reconstructed from fused features by the Decoder.

\begin{figure*}[!t]
    \centering
    \includegraphics[width=1.98 \columnwidth, ]{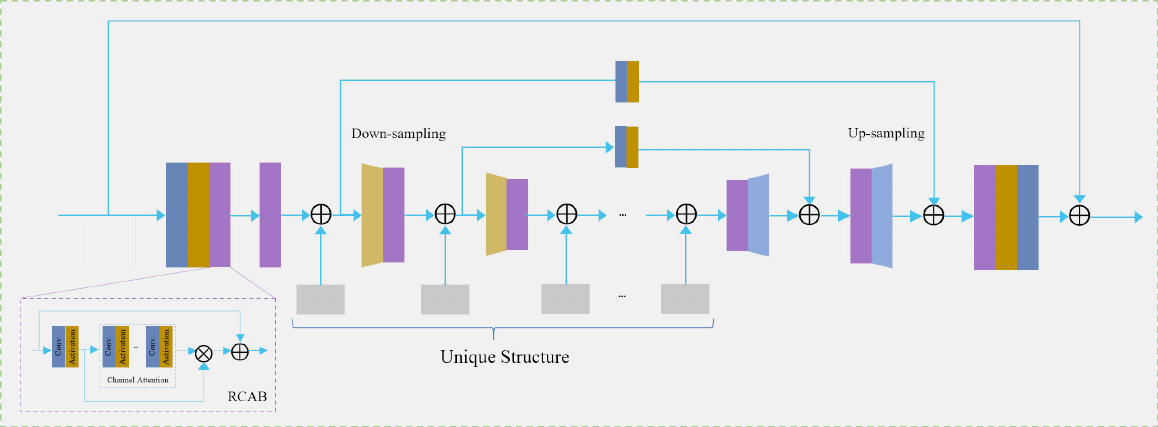}
    \caption{Architecture of feature fusion.}
    \label{fig:feature fusion}
\end{figure*}

\textbf{Fusion Details}. To facilitate effective coupling between the generated unique features and the fusion network for robust representation of cross-modal features, we made some modifications to the SFE architecture. Specifically, we introduce the residual channel attention mechanism to achieve dynamic adaptive adjustment of structural features at the channel scale, and add skip connections (or residual connections) between Encoder and Decoder to enhance the flow of structural information across different scales. 

As shown in Figure~\ref{fig:feature fusion}, the residual Channel Attention Block (RCAB)~\citep{zhang2018image} is composed of channel attention undergoing residual connections and is used for adaptive local feature perception. 
Further, the unique structure enhances these local features, facilitating the transmission of complementary structural information. 
Finally, within skipping connections, the global context information is integrated through low-level structure and high-level semantic interactions, thereby mitigating the loss of structural information due to deep convolutions.

\subsection{Training Loss}{ 
To enhance our image fusion model's sensitivity to image structures, we have devised a novel loss function outlined as:
\begin{equation} 
\mathcal{L}_{total}=\alpha \mathcal{L}_{str} + \beta \mathcal{L}_{rec} +  \mathcal{L}_{fus} ,
\label{eq}
\end{equation}
where $\alpha$ and $\beta$ control the trade-off between these three terms. We explicitly utilize a reconstruction loss $\mathcal{L}_{rec}$ to constrain structural details and prevent critical information loss in the fusion process. It can be calculated as:
\begin{equation} 
\setlength{\abovedisplayskip}{6pt}
\mathcal{L}_{rec}=\sqrt{\lVert S-S^{gt} \rVert ^2+\varepsilon ^2} ,
\setlength{\belowdisplayskip}{6pt}
\end{equation}
where $\varepsilon$ is a constant. We employ the Charbonnier Loss~\citep{Charbonnier} for $\mathcal{L}_{rec}$, which is used to estimate distance between $S$ and the ground-truth structure map~$S^{gt}$. $\mathcal{L}_{fus}$ 
encourages fused images to maintain structure consistency from the input images and can be calculated as
\begin{equation} 
\mathcal{L}_{fus}= \mathcal{L}_{ssim} + \mathcal{L}_{smooth}+ \mathcal{L}_{grad} ,
\end{equation}
where $\mathcal{L}_{ssim}$ is formulated as
\begin{equation} 
\mathcal{L}_{ssim}=1- \frac{1}{2}\left(SSIM_{ I_{fus},I_{ir}}+SSIM_{I_{fus},I_{vi}} \right)  ,
\end{equation}
where $SSIM_{A,B}$~\citep{SSIM} calculates the structural similarity of images $A$ and $B$. $\mathcal{L}_{smooth}$ and $\mathcal{L}_{grad}$ are two constraints for maintaining appearance and texture features from the input images to fused image and can be calculated as
\begin{align}    
\mathcal{L}_{smooth}=\Vert I_{fus}-max\left( I_{ir},I_{vi}\right) \Vert _1 ,
\\
\mathcal{L}_{grad}=\Vert \nabla I_{fus}-max\left( \nabla I_{ir},\nabla I_{vi} \right) \Vert _1 ,
\end{align} 
where ${\Vert } \cdot {\Vert _1}$ and $max\left(\cdot\right)$ denote the $l_1$-norm and elementwise-maximum selection operator, respectively.

\section{Experimental Results and Discussions}\label{sec:4}
A comprehensive suite of experiments has been conducted for the IVIF task to demonstrate the superior performance of our SSPFusion, with the specific experimental details outlined in Section \ref{sec:ivf}. Furthermore, experiments for the MIF task have also been tailored as presented in Section \ref{sec:mif}. The consistently outstanding results achieved across diverse scenarios underscore the robust generalization capability of our proposed method when tackling various MMIF tasks.
Finally, a discussion on the necessity of main modules in SSPFusion will be presented in Section \ref{sec:abl} through an ablation study.

%\label{sec:pagestyle}
\begin{figure*}[h]
\centering
\centerline{\includegraphics[width=2.1 \columnwidth, height=0.7\textheight]{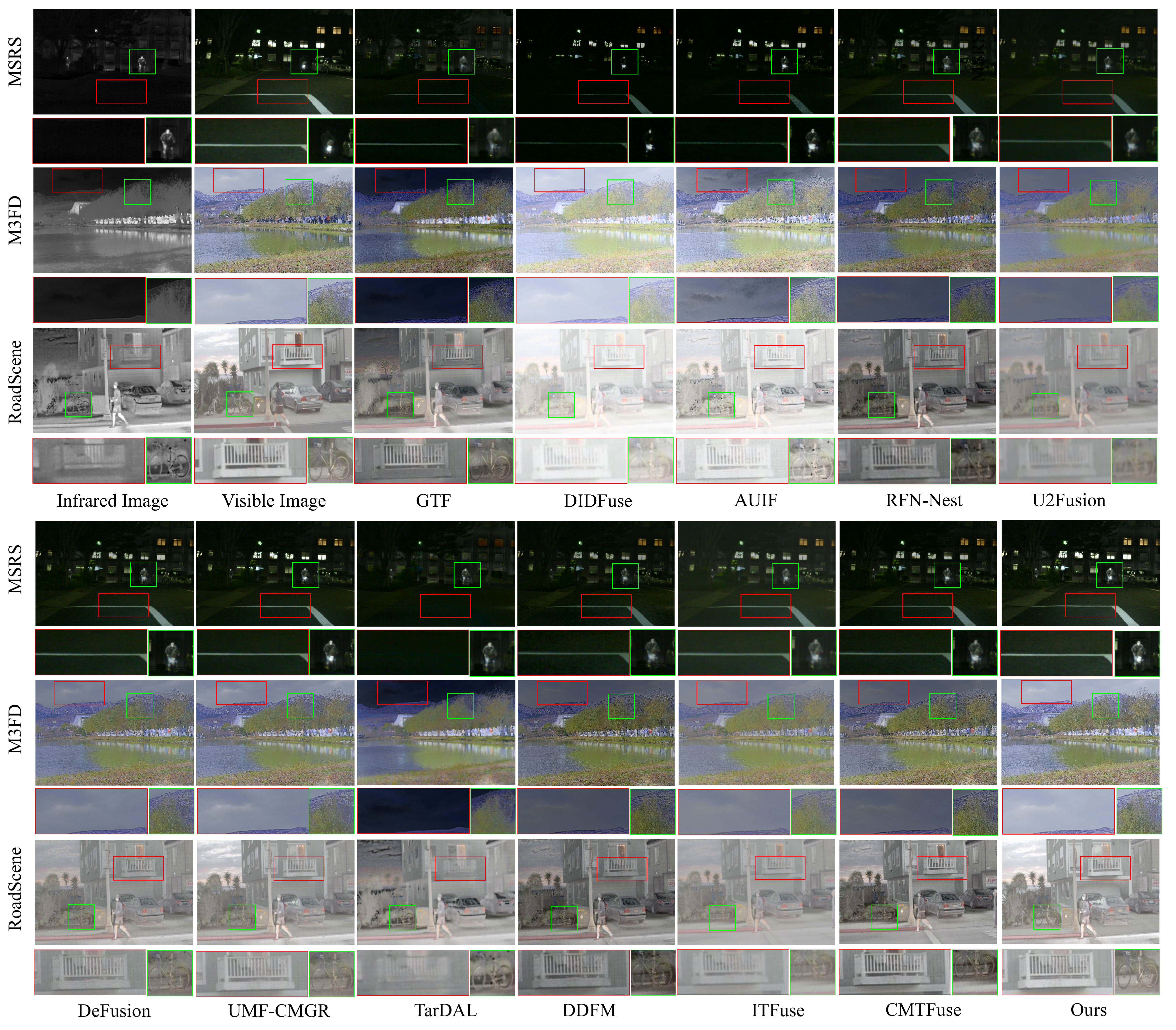}}
\caption{Qualitative comparison of different methods on fusing the ``\#00686N”, ``FLIR\_05252" and ``\#01954” pairs from the MSRS, RoadScene, M$3$FD, respectively. Close-up views of areas
within the green and red boxes were positioned in the bottom-left and bottom-right corners for better clarity in comparison.}
\label{fig:8}
\end{figure*}

\subsection{Infrared and visible image fusion}\label{sec:ivf}

We conduct experiments against nine major state-of-the-art methods, including GTF~\citep{GTF}, DIDFuse~\citep{DIDFuse}, RFN-Nest~\citep{RFN-Nest}, AUIF~\citep{AUIF}, U2Fusion~\citep{U2Fusion}, DeFusion~\citep{DeFusion}, UMF~\citep{UMF-CMGR}, TarDAL~\citep{TarDAL}, DDFM~\citep{DDFM}, ITFuse~\citep{ITFuse} and CMTFusion~\citep{CMTFusion} to verify the efficacy of our method. For a fair comparison, we fine-tune these models on the chosen dataset the MSRS dataset.
Six commonly used metrics will be employed in the experiments to ensure an all-round evaluation, including a metric measuring the amount of preserved information (MI~\citep{MI}), two metrics measuring the amount of details (SF~\citep{SF} and AG~\citep{AG}), two metrics considering human visual perception (VIF~\citep{VIF} and Qabf~\citep{Qabf}), and a metric measuring structural similarity (SSIM). 

Among the six metrics, MI quantifies the information amount of the fused image that has been preserved from multi-modality input images. SF and AG measure the textural/gradient information amount of the fused image from two different statistical views. 
Qabf utilizes local metrics to estimate the degree to which edge information from the input image is represented in the fused image.
VIF evaluates image quality based on its fidelity, which determines whether an image is visually friendly. 
SSIM comprehensively measures the correlation loss, contrast, brightness, and other aspects between the fused image and the multimodal input images. In our works, SSIM is computed here as the average of the structural similarity measurements~\citep{SSIM} between each input image and the resultant fusion image. Note that for all six metrics, a higher value indicates better quality.

\textbf{Dataset and implementation details}. 
We use three popular benchmarks, i.e., MSRS, RoadScene, and M$3$FD, to verify the efficacy of our image fusion model. 
Specifically, MSRS dataset is used to train (1083 pairs) and test (361 pairs) our model, and RoadScene (60 pairs) is used for validation. To further verify the generalization ability of our model, we directly exploit it to fuse images of the RoadScene (70 pairs) and M$3$FD (20 pairs) datasets. 
We use a two-stage training scheme inspired by~\citep{RFN-Nest}. In the stage of obtaining ground-truth structure, our model is trained on a computer platform with two NVIDIA RTX 3090 GPUs for 50 epochs, and the size of input images is set to 256$\times$256. Adam optimizer is used to optimize parameters of our model with momentum terms of $\beta_1$=0.9 and $\beta_2$=0.999. The initial learning rate is 2e-4 and decreased by the cosine annealing delay. In the fusion stage, the fused image is generated to be trained in the same way, and it is only trained for 300 epochs. In all experiments, $N$, $\alpha$, $\beta$, and $\varepsilon$ are set to $3$, $1$, $1$, and $1/1000$, respectively.

\begin{figure*}[htbp]
\centering
\centerline{\includegraphics[width=2.1 \columnwidth, height=0.48\textheight]{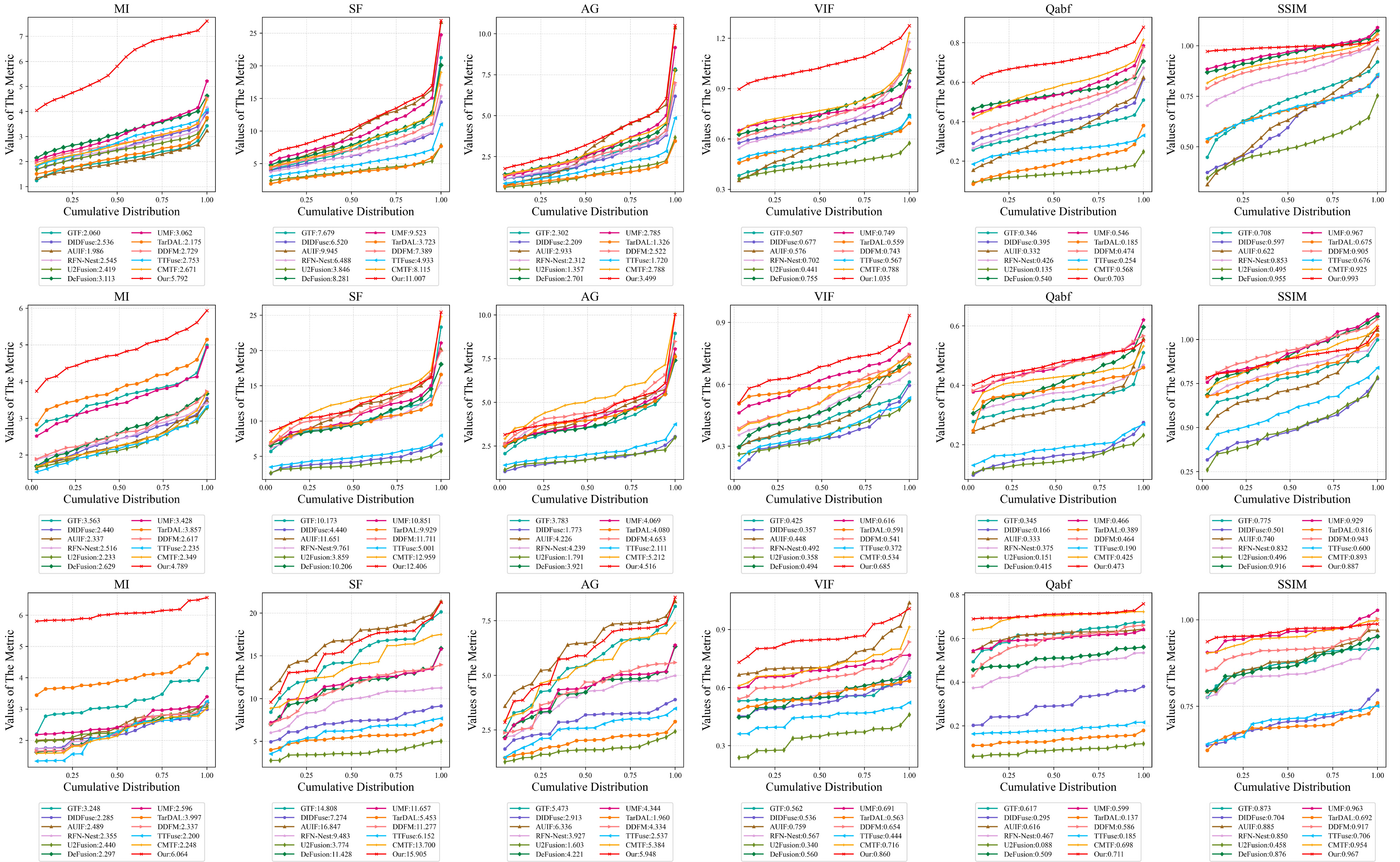}}
\caption{Quantitative comparisons of six metrics, i.e., MI, SF, AG, VIF, Qabf, and SSIM, on 361 image pairs from the MSRS dataset (the first row), 70 image pairs from the RoadScene dataset (the second row) and other 20 image pairs from the M$3$FD dataset (the third row). The nine state-of-the-art methods are used for comparison. A single point ($x,y$) on the curve denotes that there are ($100\times x$) \% percent of image pairs that have metric values no more than y.}
\label{fig:IVIFmetric}
\end{figure*}

\begin{table*}[t]
\centering
\caption{Quantitative comparison of different image fusion methods on MSRS,  RoadScene and M$3$FD dataset, values in \textbf{bold} and \underline{underlined} indicate the best and second-best results, respectively.}
\label{table1}
\resizebox{\textwidth}{!}{%
\begin{tabular}{ccccccc|cccccc|cccccc}
\toprule
     & \multicolumn{6}{c|}{MSRS dataset}                                                                            & \multicolumn{6}{c|}{RoadScene dataset}                                                                                                                            & \multicolumn{6}{c}{M$3$FD dataset}                                                                                                                                                                                                                      \\
     & MI             & SF              & AG             & VIF            & Qabf           & SSIM           & \multicolumn{1}{c}{MI} & \multicolumn{1}{c}{SF} & \multicolumn{1}{c}{AG} & \multicolumn{1}{c}{VIF} & \multicolumn{1}{c}{Qabf} & \multicolumn{1}{c|}{SSIM} & \multicolumn{1}{c}{MI}                 & \multicolumn{1}{c}{SF}               & \multicolumn{1}{c}{AG}              & \multicolumn{1}{c}{VIF}               & \multicolumn{1}{c}{Qabf}               & \multicolumn{1}{c}{SSIM}               \\ \hline
     \rule{0pt}{12pt}
GTF  & 2.060          & 7.679           & 2.302          & 0.507          & 0.346          & 0.708          & 3.563                  & 10.173                 & 3.783                  & 0.425                   & 0.345                    & 0.775                     & 3.248                                  & 14.808                               & 5.473                               & 0.562                                 & 0.617                                  & 0.873                                  \\
DID  & 2.536          & 6.520           & 2.209          & 0.677          & 0.395          & 0.597          & 2.440                  & 4.440                  & 1.773                  & 0.357                   & 0.166                    & 0.501                     & 2.285                                  & 7.274                                & 2.913                               & 0.536                                 & 0.295                                  & 0.704                                  \\
AUIF & 1.986          & {\ul 9.945}     & {\ul 2.933}    & 0.576          & 0.332          & 0.622          & 2.337                  & 11.651                 & 4.226                  & 0.448                   & 0.333                    & 0.740                     & 2.489                                  & \textbf{16.847}                      & \textbf{6.336}                      & {\ul 0.759}                           & 0.616                                  & 0.885                                  \\
RFN  & 2.545          & 6.488           & 2.312          & 0.702          & 0.426          & 0.853          & 2.516                  & 9.761                  & 4.239           & 0.492                   & 0.375                    & 0.832                     & 2.355                                  & 9.483                                & 3.927                               & 0.567                                 & 0.467                                  & 0.850                                   \\
U2F  & 2.419          & 3.846           & 1.357          & 0.441          & 0.135          & 0.495          & 2.233                  & 3.859                  & 1.791                  & 0.358                   & 0.151                    & 0.496                     & 2.440                                  & 3.774                                & 1.603                               & 0.340                                  & 0.088                                  & 0.458                                  \\
DeF  & {\ul 3.113}    & 8.281           & 2.701          &  0.755    & 0.540           & 0.955          & 2.629                  & 10.206                 & 3.921                  & 0.494                   & 0.415                    & 0.916                     & 2.297                                  & 11.428                               & 4.221                               & 0.560                                  & 0.509                                  & 0.876                                  \\
UMF  & 3.062          & 9.523           & 2.785          & 0.749          & 0.546    & {\ul 0.967}    & 3.428                  & 10.851                 & 4.069                  & {\ul 0.616}             & {\ul 0.466}              & {\ul 0.929}               & 2.596                                  & 11.657                               & 4.344                               & 0.691                                 & 0.599                                  & {\ul 0.963}                            \\
TarD & 2.175          & 3.723           & 1.326          & 0.559          & 0.185          & 0.675          & {\ul 3.857}            & 9.929                  & 4.080                   & 0.591                   & 0.389                    & 0.816                     & 3.997                                  & 5.453                                & 1.960                                & 0.563                                 & 0.137                                  & 0.692                                  \\
DDFM & 2.729          & 7.389           & 2.522          & 0.743          & 0.474          & 0.905          & 2.617                  & 11.711           & 4.653         & 0.541                   & 0.464                    & \textbf{0.943}            & 2.337                                  & 11.277                               & 4.334                               & 0.654                                 & 0.586                                  & 0.917                                  \\
% LLRNet & 2.922          & 8.471           & 2.651          & 0.541          & 0.454          & 0.422          & 2.563                  & 12.943           & {\ul 4.898}         & 0.465                   & 0.359                    & 0.674            & 2.533                                  & 12.672                               & 4.473                               & 0.605                                 & 0.606                                  & 0.837                                  \\
ITFuse & 2.753&4.933&1.720&0.567&0.254&0.676&2.235&5.001&2.111&0.372&0.190&0.600&2.200&6.152&2.537&0.444&0.185&0.706                                  \\
CMTF & 
2.671&8.115&2.788&{\ul0.788}&{\ul 0.568}&0.925&2.349&{\ul 12.959}&\textbf{5.212}&0.534&0.425&0.893&2.248&13.700&5.384&0.716&0.698&0.954                                  \\
\cellcolor[HTML]{ECF4FF}
Ours & \cellcolor[HTML]{ECF4FF}\textbf{5.792} & \cellcolor[HTML]{ECF4FF}\textbf{11.007} & \cellcolor[HTML]{ECF4FF}\textbf{3.499} & \cellcolor[HTML]{ECF4FF}\textbf{1.035} & \cellcolor[HTML]{ECF4FF}\textbf{0.703} &\cellcolor[HTML]{ECF4FF} \textbf{0.993} & \cellcolor[HTML]{ECF4FF}\textbf{4.789}         & \cellcolor[HTML]{ECF4FF}\textbf{12.406}        & \cellcolor[HTML]{ECF4FF} 4.516            &\cellcolor[HTML]{ECF4FF} \textbf{0.685}          & \cellcolor[HTML]{ECF4FF}\textbf{0.473}           & \cellcolor[HTML]{ECF4FF}0.887                     & \cellcolor[HTML]{ECF4FF}\textbf{6.064} & \cellcolor[HTML]{ECF4FF}{\ul 15.905} & \cellcolor[HTML]{ECF4FF}{\ul 5.948} & \cellcolor[HTML]{ECF4FF}\textbf{0.860} & \cellcolor[HTML]{ECF4FF}\textbf{0.711} & \cellcolor[HTML]{ECF4FF}\textbf{0.967} \\ 
\bottomrule
\end{tabular}%
}
\label{tablexxx}
\end{table*}

\textbf{Performance analysis}. 
In this part, we thoroughly analyze the experimental results of our proposed SSPFusion on the IVIF task.

\textit{Qualitative evaluation.} Three representative samples from various image scenarios are selected as shown in Figure~\ref{fig:8} to demonstrate that our SSPFusion can outperform other methods, particularly in terms of structure preservation, complement information, and color fidelity.
The sample ``\#00686N" from MSRS dataset illustrates an urban night scenario. Due to its complex illumination conditions, the major challenges for the IVIF task lie in the road markings by the red boxes, as well as the cyclists by the green boxes. 
The state-of-the-art methods either fail to recover the objects, i.e. the cyclists in DIDFuse, the road marking in TarDAL, or suffer an edge-blurring effect, i.e. the blurring cyclists in GTF, RFN-Nest, U2Fusion, DDFM, and CMTFusion. While preserving both the structural information and texture details, our SSPFusion achieves a closer result in color fidelity to the visible image than DeFusion and UMF. The sample ``FLIR\_05252" from RoadScene dataset illustrates an overexposed road landscape scenario. Despite their decent performance in restoring the road markings and cyclists in the previous example, both DeFusion and UMF exhibit poor restoration of the bicycle within the green boxes in this example. Attention should be paid that when other methods suffer from noticeable color distortion under the overexposure condition, our method is still able to achieve a relatively reasonable color representation. As for the sample ``\#09154" from M$3$FD dataset, our method continues to achieve the most authentic color restoration while preserving the texture of the clouds within the red boxes and the details of the trees within the green boxes. 
In the fusion results of GTF, RFN-Nest, U2Fusion, TarDAL, DDFM, ITFuse, and CMTFusion, there is a significant deviation in the color of the sky. Additionally, while DIDFuse presents a normal-looking sky, other image elements like trees and water exhibit an overexposed quality. In contrast, DeFusion and UMF depict trees and water accurately, but the sky appears underexposed. Ultimately, our method stands out as the only one capable of simultaneously and precisely restoring both the intricate details and authentic colors of both the sky and trees.

\begin{figure*}[h]
\centering
\centerline{\includegraphics[width=2.0 \columnwidth]{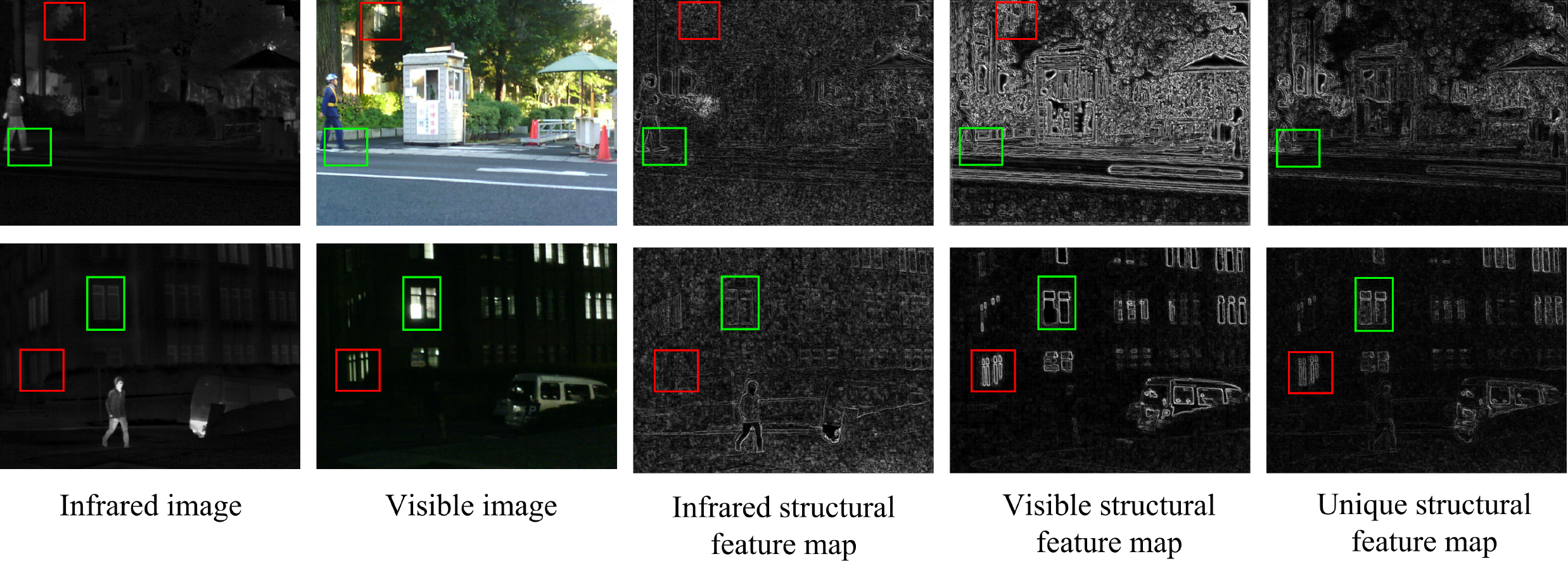}}
\caption{Visualization of the unique structure on fusing the ``\#00097D” and ``\#00054N" pairs from the MSRS.}
\label{fig:unique}
\end{figure*}

\textit{Quantitative evaluation.}{
As shown in Figure~\ref{fig:IVIFmetric}, our SSPFusion has a clear lead in MI, VIF, Qabf, and SSIM, indicating our fused images get higher information gain and maintain better structural consistency against the input images. Additionally, our method also achieves competitive results on other metrics. In terms of the metrics SF and AG, our method is second to AUIF only in M$3$FD dataset, but in other datasets and on other metrics, our SSPFusion is significantly ahead. 
This is also consistent with the visualization results presented in Figure~\ref{fig:8}, where our SSPFusion also outperforms other methods.
Meanwhile, we consolidate the numerical results of six metrics across three datasets in Table~\ref{tablexxx} to facilitate a clear and straightforward comparison of the values.}

\textbf{Visualization of unique structure.} In Figure~\ref{fig:unique}, we visually illustrate the unique structural characteristics that emerge during the transfer of cross-modal information through the visualization of intermediate layer features.

As shown in the 1st and 2nd columns of sample ``\#00097D" in the first row, both infrared and visible images exhibit distinct edge structures of pedestrians in good illumination conditions. However, when those edge structure features propagate through the network, the human foot within the green boxes becomes submerged in the background in the visible structure feature map, as well as the building details in the red boxes. 
On the contrary, the sample ``\#00054N" in the second row in Figure~\ref{fig:unique} demonstrates a similar scenario in bad illumination conditions. At this time, the windows from the building in green and red boxes are relatively blurry in the infrared feature map, while clear in the visible feature map. 

Thanks to the application of the proposed $\mathcal{J}_p(\cdot)$ function, the generated unique structural feature map contains complementary information, which enables these parts in the boxes to be effectively detected.
Thus, we can utilize these unique structural feature maps to enhance the original ones, resulting in semantic features that enable robust cross-modal representation in all kinds of illumination conditions.

\textbf{Performance on semantic segmentation task.} To explore the influence of our method on high-level vision tasks, we test its performance for semantic segmentation on the MSRS dataset. 
Specifically, we select DeeplabV3+~\citep{DeeplabV3+} as the baseline segmentation model, train the segmentation model on the MSRS dataset (trained on 976 images and tested on 361 images), and evaluate the model's performance by Intersection over Union (IoU) on 9 object categories: Background (BaC), Car, Person (Per), Bike (Bik), Curve (Cur), Car Stop (CS), Guardrail (GD), Color Cone (CC), Bump (Bu), and their mean IoU (mIoU).

\begin{figure*}[!t]
\centering
\centerline{\includegraphics[width=2.1 \columnwidth, height=0.6\textheight]{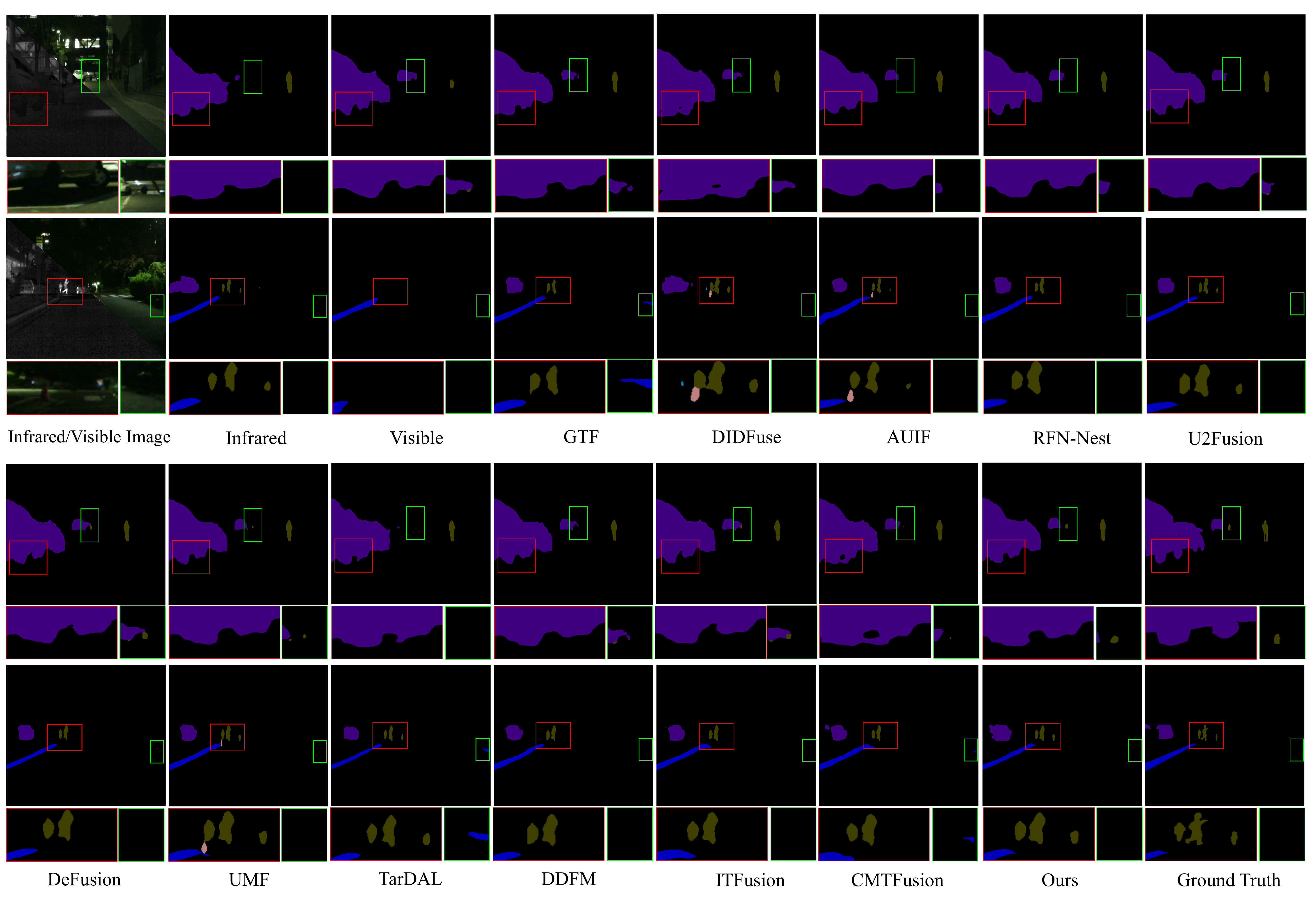}}
\caption{Segmentation results of the fusing the ``\#01053N” and ``\#00878N" pairs from the MSRS. For clear comparisons, the areas boxed in green and red are shown specifically in the fused image. Close-up views of areas
within the green and red boxes were positioned in the bottom-left and bottom-right corners for better clarity in comparison.
}
\label{fig:11}
\end{figure*}

\begin{table*}[h]
\centering

\caption{Comparison of segmentation results on MSRS dataset, values in \textbf{bold} indicate the best and second-best results, respectively.}
\label{tab:segmentation}
\renewcommand{\arraystretch}{1.2}
\begin{tabular}{l|l|l|l|l|l|l|l|l|l|l}
\hline
{Method } & Bac & Car & Per & Bik & Cur & CS & GD & CC & Bu & mIoU \\ \hline 
Infrared &97.6 	&83.6 	&70.4 	&64.7 	&50.5 	&55.4 	&50.6 	&48.5 	&59.8 	&64.6 \\
Visible & 97.4 & 83.2 & 52.8 & 64.5 & 43.3 & 58.9 & 72.1 & 53.7 & 60.0 & 65.2 \\
GTF~\citep{GTF} & 97.6 & 83.8 & 67.7 & 66.1 & 45.0 & 61.3 & 61.3 & 48.1 & 58.5 & 65.5 \\
DID~\citep{DIDFuse} & 97.6 & 84.5 & 63.7 & 65.6 & 45.3 & 61.4 & 71.8 & 51.1 & 56.1 & 66.3 \\
AUIF~\citep{AUIF} & 97.9 & 86.2 & 67.3 & 68.3 & 49.1 & 65.6 & 77.9 & 55.1 & 56.5 & 69.3 \\
RFN~\citep{RFN-Nest} & 98.1 & 87.1 & 68.7 & 69.2 & 55.3 & \textbf{67.8} & 76.7 & 57.1 & 66.0 & 71.8 \\
U2F~\citep{U2Fusion} & 98.0 & 86.8 & 67.9 & 68.1 & 54.3 & 66.5 & 76.4 & 56.8 & 67.2 & 71.3 \\
DeF~\citep{DeFusion} & 98.1 & 86.9 & 68.9 & 68.6 & 54.0 & 66.4 & 77.6 & 56.2 & 66.2 & 71.4 \\
UMF~\citep{UMF-CMGR} & 98.1 & 86.9 & 67.8 & 67.9 & 54.7 & 67.1 & 78.4 & 56.9 & 67.4 & 71.7 \\
TarD~\citep{TarDAL} & 97.9 & 86.2 & \textbf{71.5} & 67.1 & 52.6 & 62.6 & 66.0 & 55.7 & 58.6 & 68.7 \\
DDFM~\citep{DDFM} & 98.1 & 87.1 & 68.4 & 68.8 & 53.4 & 67.3 & 76.0 & 56.1 & 63.8 & 71.0 \\
ITFuse~\citep{ITFuse} & 98.0 & 86.7 & 68.1 & 68.1 & 53.3 & 66.2 & 75.9 & 57.1 & 61.6 & 70.6 \\
CMT~\citep{CMTFusion} & 98.1 & 86.8 & 68.6 & \textbf{69.4} & 55.5 & 65.6 & 74.3 & 57.2 & 67.4 & 71.5 \\
{Ours} & \textbf{98.1} & \textbf{87.2} & {68.4} & {68.4} & \textbf{56.7} & {65.7} & \textbf{78.6} & \textbf{57.3} & \textbf{70.0} & \textbf{72.2} \\ \hline
\end{tabular}%
%}
%}
\end{table*}

\textit{Implementation Details.} For DeeplabV3+, all the models were trained with both Dice and Cross-entropy loss, and optimized by SGD over 340 epochs with a batch size of 4. The first 100 epochs were trained by freezing the baseline with a batch size of 8. The initial learning rate was 7e-3 and decreased by the cosine annealing delay.

\textit{Qualitative evaluation.} Two representative samples are selected to demonstrate the segmentation results as shown in Figure~\ref{fig:11}. Our SSPFusion performs doubtlessly the best among the six state-of-the-art methods. In sample ``\#01053N", most of the other methods fail to correctly segment the distant pedestrian in the green box, except for DeFusion, and UMF. Furthermore, our method achieves the highest segmentation quality of both the pedestrian and the nearside vehicle. In sample ``\#00878N", our SSPFusion and U2Fusion stand out as the remaining methods that are capable of correctly segmenting all three pedestrians and curbs in the road within the red box, while simultaneously avoiding generating erroneous noise within the green box. Taking a comprehensive view across multiple samples, our method undoubtedly ranks among the top in terms of both quality and stability in segmentation. 

\textit{Quantitative Comparison.} 
The results in Table~\ref{tab:segmentation} indicates our method achieves the highest IoU for the four categories and outperforms other methods on the semantic segmentation task. Specifically, our method achieves the best performance in the categories of Curve, Guardrail, Color Cone, and Bump, demonstrating its superior capability in feature extraction for small targets (e.g., Color Cone), low-texture targets (e.g., Guardrail), and geometry-sensitive structures (e.g., Curve and Bump). Notably, our method obtains a mIoU of 72.2\%, significantly surpassing all competing methods, which highlights its strong competitiveness in terms of overall segmentation accuracy. In particular, the substantial improvements over single-modal inputs (64.6\% for infrared images and 65.2\% for visible images) further confirm the necessity and effectiveness of multi-modal fusion.

Both quantitative and qualitative comparisons fully validate the superior potential of the proposed approach over state-of-the-art image fusion algorithms for high-level vision tasks.

\subsection{Medical image fusion}
\label{sec:mif}
To further validate the good generalization ability of our method, we compare it with three state-of-the-art medical or general image fusion methods, i.e. IFCNN~\citep{IFCNN}, EMFusion~\citep{EMFusion}, DDFM~\citep{DDFM}, MATR~\citep{MATR}, ALMF~\citep{ALMF}, and TUFusion~\citep{TUFusion}. We have selected the same six evaluation metrics used in the IVIF task to assess our model.

\textbf{Dataset and implementation details.}
To test the performance of our model on the medical image fusion task, we collected 357 pairs of MRI and SPECT images from the public Harvard medical image dataset, in which 257, 40, and 60 pairs were used for training, validation, and testing, respectively. 
Images was randomly cropped to $64 \times 64$ before feeding to our model. The training strategies were the same as those on the IVIF task.

\begin{figure}[!t]
\centering
\centerline{\includegraphics[width=0.95\columnwidth, height=0.5\textheight]{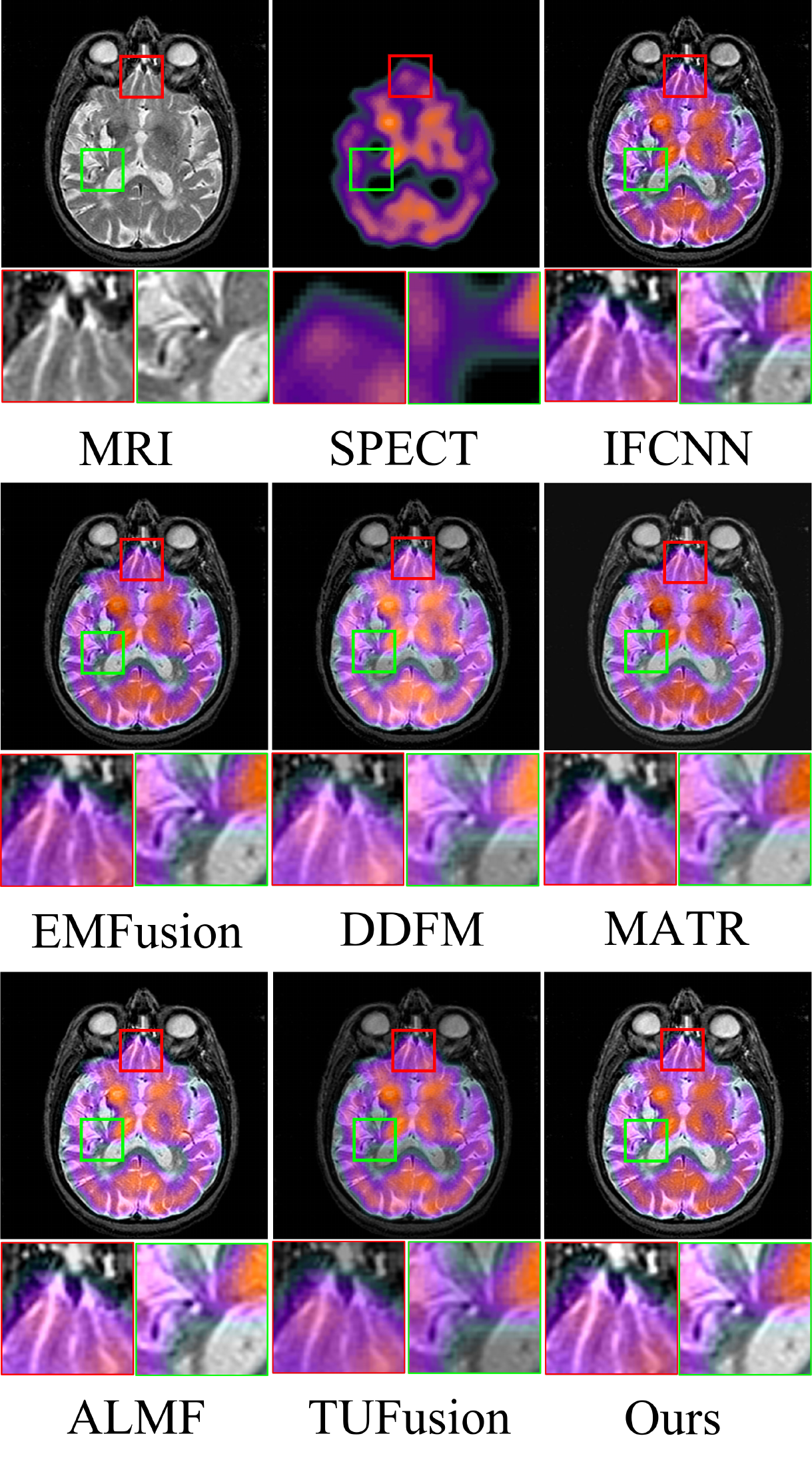}}
\caption{Qualitative comparison of different methods on fusing the ``\#18023” pairs from the SPECT-MRI dataset. Close-up views of areas within the green and red boxes were positioned in the bottom-left and bottom-right corners for better clarity in comparison.}
\label{fig:medical}
\end{figure}

\begin{figure*}[h]
\centering
\centerline{\includegraphics[width=2.1 \columnwidth]{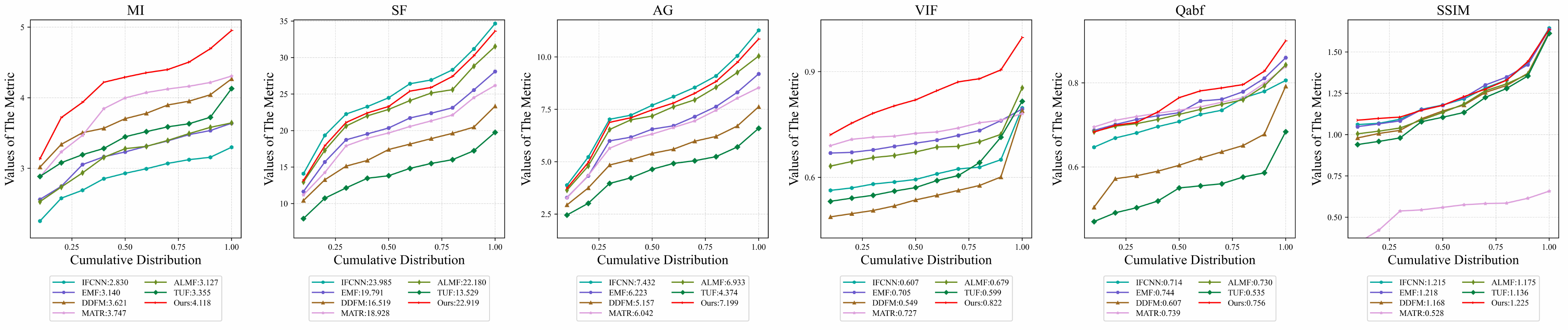}}
\caption{Quantitative comparisons of six metrics, i.e., MI, SF, AG, VIF, Qabf, and SSIM, on 60 image pairs from the SPECT-MRI dataset. A single point ($x,y$) on the curve denotes that there are ($100\times x$) \% percent of image pairs that have metric values no more than y.}
\label{fig:13}
\end{figure*}

\begin{table}[h]
%\vspace{-10mm}
\centering % 表格居中
\caption{Quantitative evaluation results on the SPECT-MRI dataset, values in \textbf{bold} and \underline{underlined} indicate the best and second-best results, respectively.}
%\resizebox{0.50\linewidth}{!} % 设计表格尺寸
%\setlength{\tabcolsep}{1mm}
\renewcommand{\arraystretch}{1.2}
\begin{tabular}{l|l|l|l|l|l|l} % 每列设计 
\hline %第一根加粗线
   Method  & MI  & SF  & AG  & VIF  & Qabf  & SSIM   \\ \hline
  IFCNN    & 2.830  & \textbf{23.985}  & \textbf{7.432}   & 0.607    & 0.714     & 1.215 \\
  EMF      & 3.140  & 19.791    & 6.223  & 0.705   & \underline{0.744}    & \underline{1.218}  \\
      DDFM    & 3.621     & 16.519    & 5.157          & 0.549         & 0.607          & 1.168      \\
  MATR    & \underline{3.747}     & 18.928    & 6.042          & \underline{0.727}         & 0.739          & 0.528      \\
    ALMF    & 3.127     & 22.180    & 6.933          & 0.679         & 0.730          & 1.175      \\
        TUF    &3.355     & 13.529    & 4.374          & 0.559         & 0.535          & 1.136      \\
  \cellcolor[HTML]{ECF4FF}Ours     & \cellcolor[HTML]{ECF4FF}\textbf{4.118}       & \cellcolor[HTML]{ECF4FF}\underline{22.919}    & \cellcolor[HTML]{ECF4FF}\underline{7.199}         & \cellcolor[HTML]{ECF4FF}\textbf{0.822}    & \cellcolor[HTML]{ECF4FF}\textbf{0.756}   & \cellcolor[HTML]{ECF4FF}\textbf{1.225}\\ \hline %第一根加粗线
\end{tabular}
%}
\label{table3}
%}
\end{table}

\textbf{Performance analysis.} 
In this part, we further analyze the experimental results of our proposed SSPFusion on the MIF task.

\textit{Qualitative evaluation.} One comparison example has been shown in Figure~\ref{fig:medical}, in which SPECT image mainly shows the blood-flow or metabolism status of the lesion area and MRI image contains of more textural details. 
Considering the MRI textures within the magnified green box in Figure~\ref{fig:medical}, our fused image demonstrates the most impressive restoration effect. In contrast, results from EMFusion and DDFM exhibit a distortion on either overall color fidelity or local texture. In summary, our method excels in retaining more significant features from the input images in its fused images compared to other methods.

\textit{Quantitative evaluation.}{We also use the above six quantitative metrics in the MIF task. As shown in Figure~\ref{fig:13}, our SSPFusion has also a clear lead in MI, VIF, Qabf, and SSIM, indicating our fused images get higher information gain and maintain better structural consistency against the SPECT-MRI images. Additionally, our method also ranks second in terms of SF and AG metrics. As shown in Table~\ref{table3}, our method is significantly better than the second-best method in MI and VIF metrics, indicating our method can transfer more structural textures while suppressing redundant information to its fused images. Additionally, our method achieves competitive results on other metrics, similar to those observed in the evaluation of the IVIF task, which further proves the capability of SSPFusion in various MMIF tasks.
}

\begin{table}[t]
	\begin{center}
\caption{Ablation study of $\mathcal{F}$ and $\mathcal{P}$ on MSRS and SPECT-MRI datasets. Values in \textbf{bold} indicate the best results.}
\label{table4}
\begin{tabular}{cc|cccccc}
\toprule
$\mathcal{F}$ & $\mathcal{P}$ & MI & SF & AG & VIF & Qabf & SSIM \\
\midrule
\multicolumn{8}{c}{\textbf{MSRS}} \\
\midrule
\ding{55} & \ding{51} & 5.523 & 10.902 & 3.446 & 0.978 & 0.652 & \textbf{1.028} \\
\ding{51} & \ding{55} & 5.393 & 10.846 & 3.424 & 0.961 & 0.637 & 0.965 \\
\ding{51} & \ding{51}& \textbf{5.792} & \textbf{11.007} & \textbf{3.499} & \textbf{1.035} & \textbf{0.703} & 0.993 \\
\midrule
\multicolumn{8}{c}{\textbf{SPECT-MRI}} \\
\midrule
\ding{55} & \ding{51} &3.128 & 18.911 & 4.432 & 0.663 & 0.721 & 1.147 \\
\ding{51} & \ding{55} & 3.029 & 22.649 & 6.962 & 0.698 & 0.666 & 1.005 \\
\ding{51} & \ding{51} & \textbf{4.118} & \textbf{22.919} & \textbf{7.199} & \textbf{0.822} & \textbf{0.756} & \textbf{1.225} \\
\bottomrule
\end{tabular}
\end{center}
\end{table}

\subsection{Ablation analysis}\label{sec:abl}
To analyze the effectiveness of SSPFusion, we conduct ablation studies to reveal the influence of key components in our method. 
Ablation study on $\mathcal{F}$ and $\mathcal{P}$ is provided in Table~\ref{table4}. Specifically, we design ablation experiments to verify the effectiveness of SFE and SPF on MSRS dataset. Among them, $\mathcal{F}$ represents SFE and $\mathcal{P}$ represents SPF.
We compare the performance of our method with/without (\ding{51} and \ding{55}) them, respectively. 
For the SFE module, the training loss was revised to exclude $\mathcal{L}_{str}$ and $\mathcal{L}_{rec}$, focusing solely on $\mathcal{L}_{fus}$. Similarly, the SPF module involved the direct removal of this particular component from the training process.
As shown in Table~\ref{table4}, when incorporating SFE and SPF, our SSPFusion can integrate more salient and complementary features of infrared and visible images into their fusion images compared to solely leveraging either SFE or SPF individually. 
As further illustrated in Figure~\ref{fig:14}, when both $\mathcal{F}$ and $\mathcal{P}$ are present, the generated images successfully preserve the structural information from either visible image or MRI image while simultaneously incorporating the salient targets from infrared imgae or SPECT image.

Moreover, Figure~\ref{fig:15} presents the feature maps under different configurations, including with/without module $\mathcal{F}$ and with/without module $\mathcal{P}$. Generally, in visualized feature maps, darker shades of red (approaching 1.0) indicate higher importance/activation, while darker shades of blue (approaching 0.0) indicate lower importance/activation. As shown in Figure 15, the absence of module $\mathcal{F}$ (as illustrated in (b) and (d)) leads to feature representations overemphasizing the global characteristics while neglecting critical semantic relationships in local regions. Similarly, the absence of module $\mathcal{P}$ (as illustrated in (f)) causes the method to fail to fully extract prominent structural cues necessary for effective fusion. The inclusion of module $\mathcal{F}$ (as illustrated in (c) and (e)) enables the network to retain both low-frequency structural and high-frequency semantic information, resulting in clearer and more continuous feature representations. Meanwhile, module $\mathcal{P}$ (as illustrated in (g)) enhances the structural integrity of the fused features by leveraging unique structural cues. As shown in the last subfigure in the bottom-right corner, the synergistic interaction between modules $\mathcal{F}$ and $\mathcal{P}$ significantly enhances the discriminative power and completeness of the extracted features when both are present.

\begin{figure}[h]
\centering
\centerline{\includegraphics[width=0.98 \columnwidth]{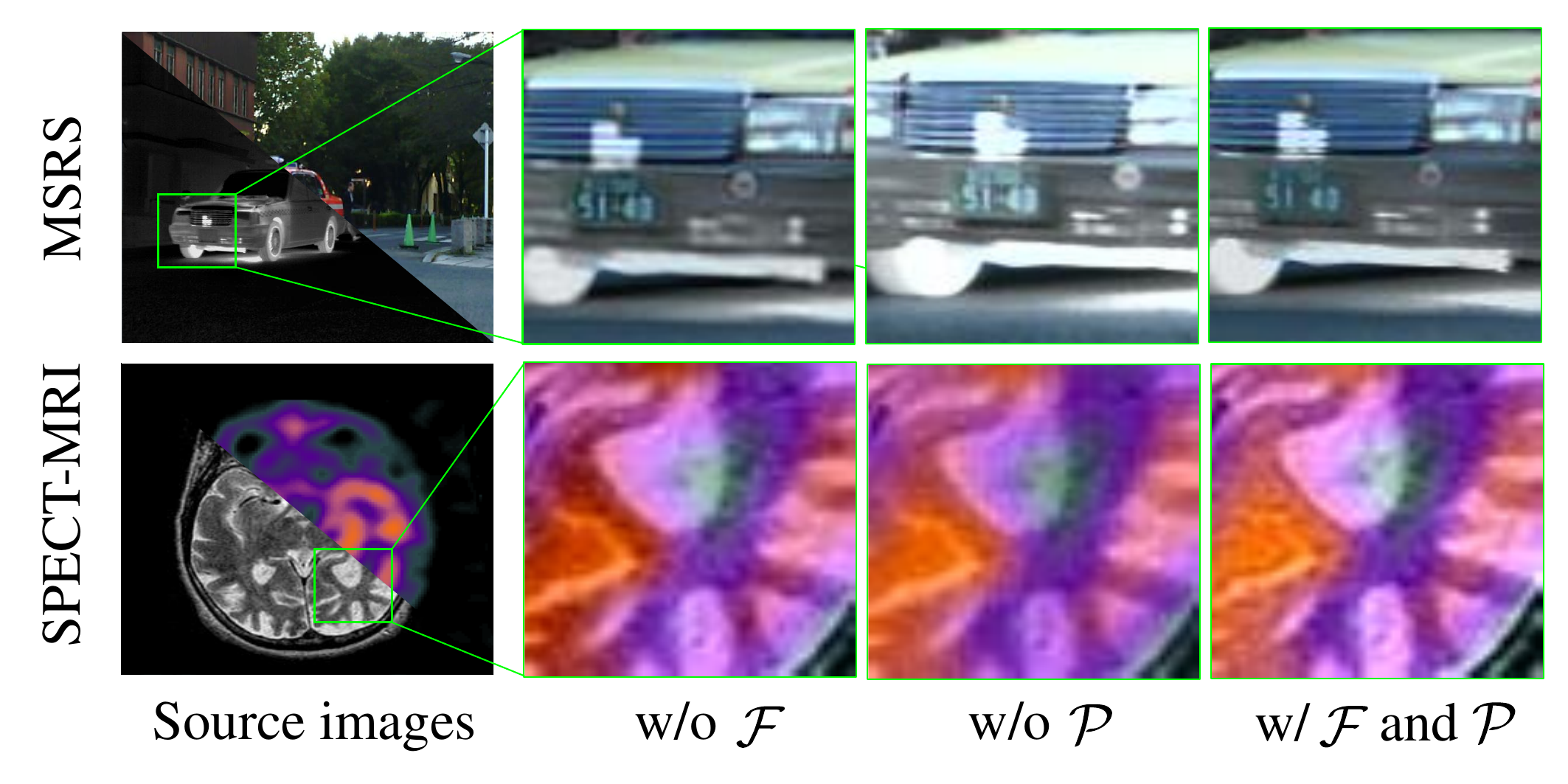}}
\caption{Progressive fusion results on the MSRS (1th row), and SPECT-MRI (2th row). From left to right: source images, w/o $\mathcal{F}$, w/o $\mathcal{P}$, and the full model of ours.}
\label{fig:14}
\end{figure}

\begin{figure}[h]
\centering
\centerline{\includegraphics[width=1.02 \columnwidth]{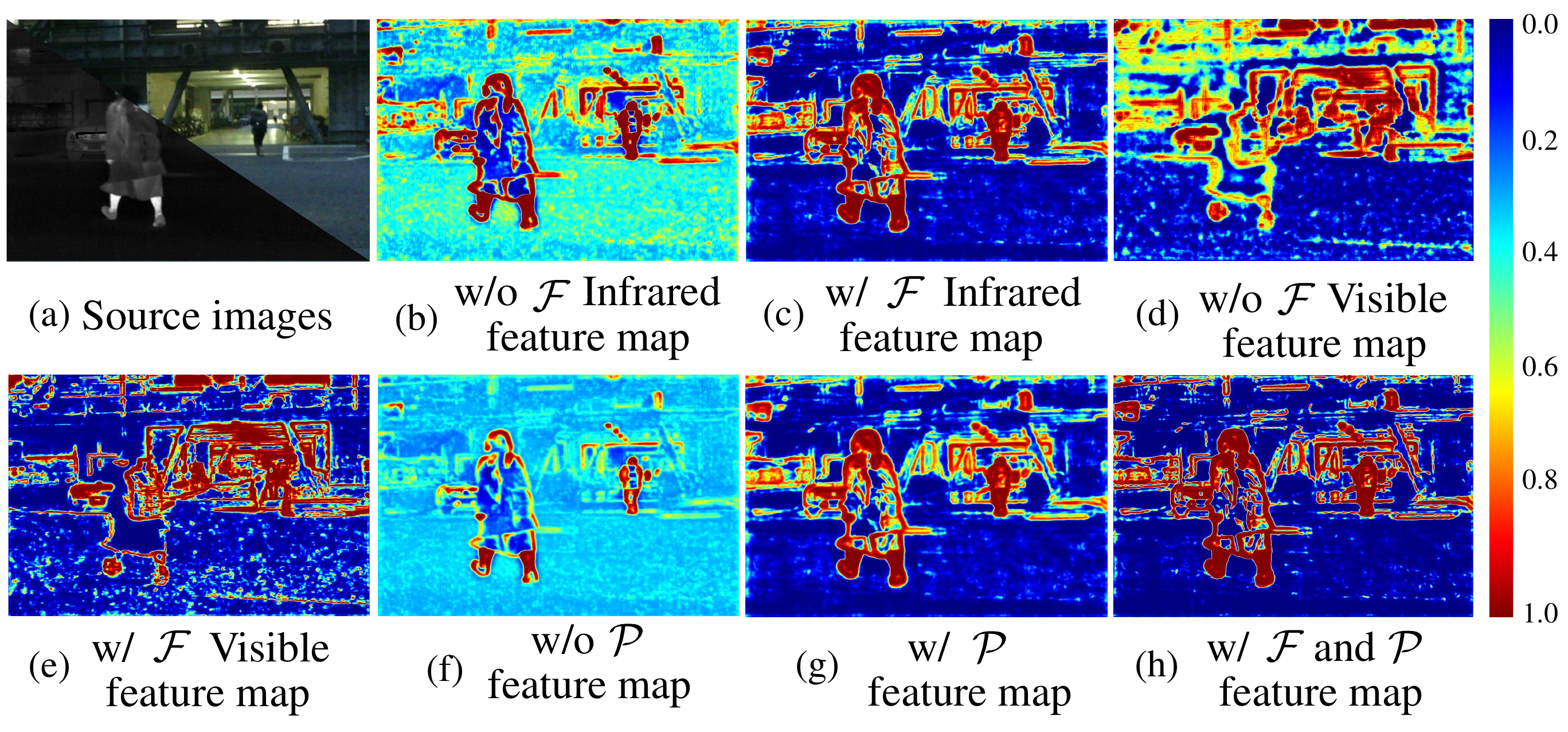}}
\caption{The visualization of the extracted typical feature maps from different modality images. Each image is accompanied by its corresponding description below.}
\label{fig:15}
\end{figure}

%\label{sec:pagestyle}
\begin{figure*}[h]
\centering
\centerline{\includegraphics[width=2.1 \columnwidth]{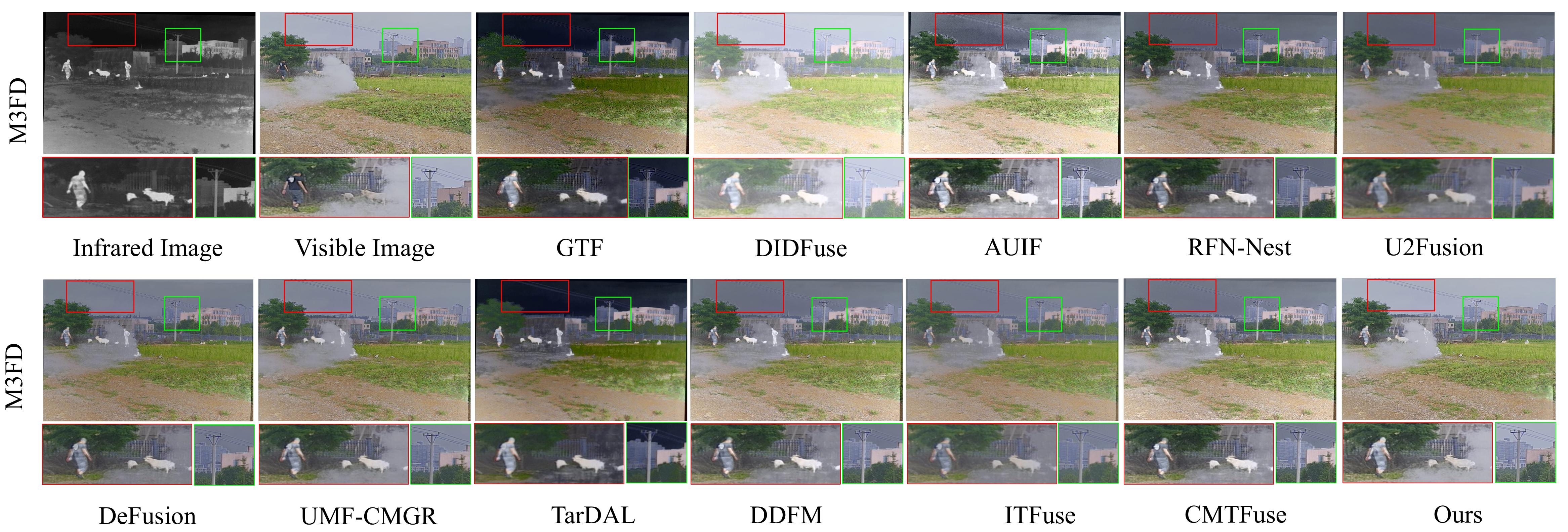}}
\caption{Qualitative comparison of different methods on fusing the ``\#00878” pairs from the M$3$FD. Close-up views of areas within the green and red boxes were positioned in the bottom-left and bottom-right corners for better clarity in comparison.
}
\label{fig:16}
\end{figure*}

%\label{sec:pagestyle}
\begin{figure*}[h]
\centering
\centerline{\includegraphics[width=2.1 \columnwidth]{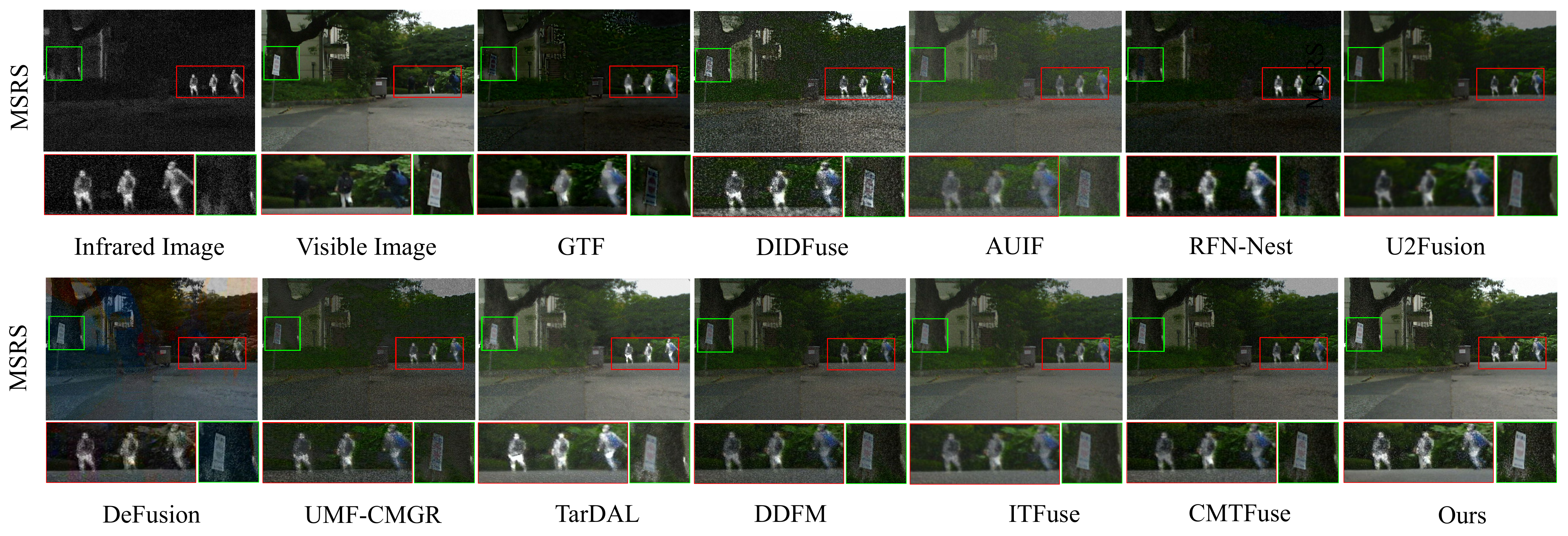}}
\caption{Qualitative comparison of different methods on fusing the ``\#00638D” pairs from the MSRS. Close-up views of areas within the green and red boxes were positioned in the bottom-left and bottom-right corners for better clarity in comparison.}
\label{fig:17}
\end{figure*}

%\label{sec:pagestyle}
\begin{figure}[h]
\centering
\centerline{\includegraphics[width=1.1\columnwidth]{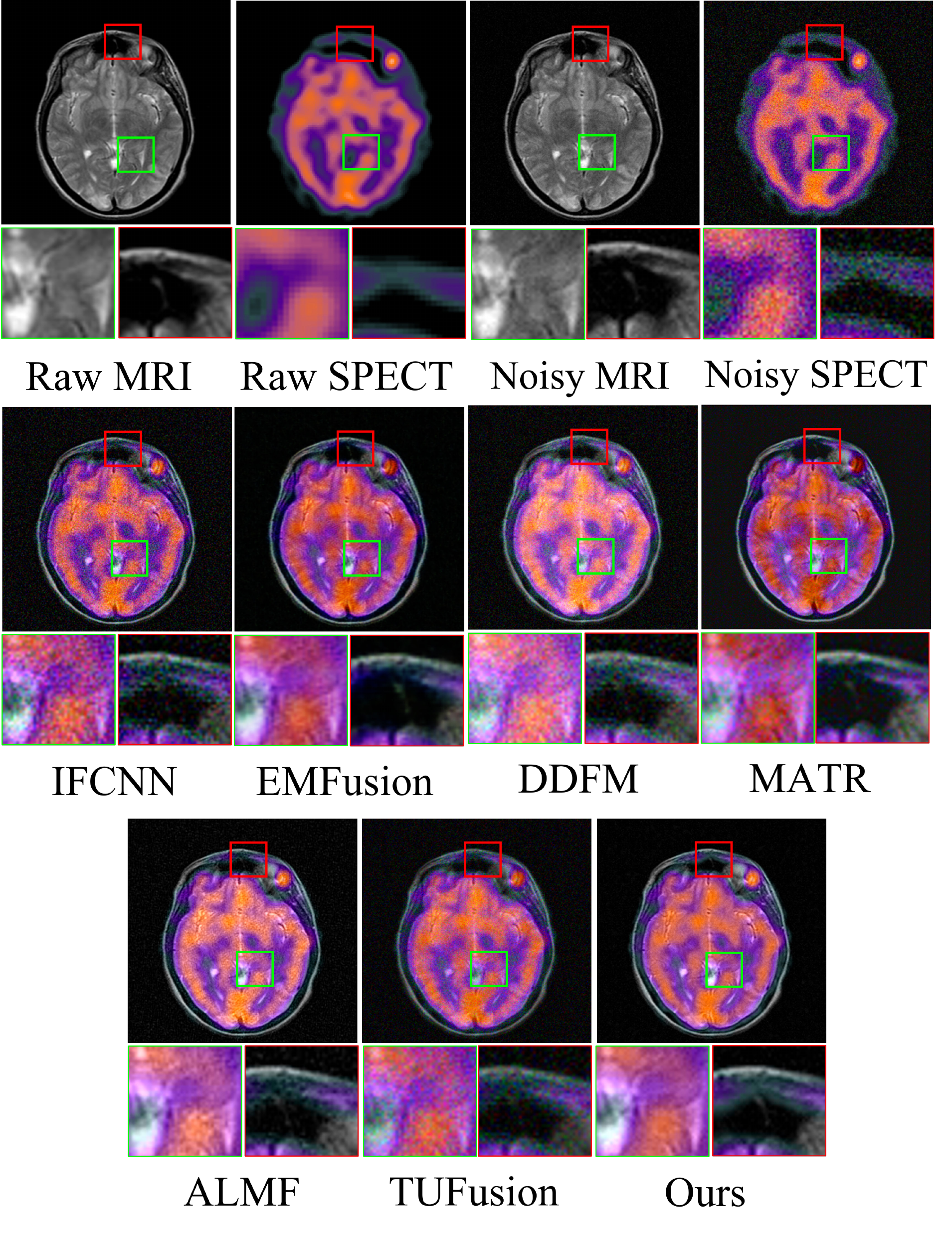}}
\caption{Qualitative comparison of different methods on fusing the ``\#21011” pairs from the SPECT-MRI. Close-up views of areas within the green and red boxes were positioned in the bottom-left and bottom-right corners for better clarity in comparison.}
\label{fig:18}
\end{figure}

\subsection{Discussions}
\textbf{Performance on Occlusion/Noise Scenario.} The proposed method demonstrates remarkable advantages in standard scenarios and achieves state-of-the-art performance among existing models, as evidenced by the results in Figures~\ref{fig:8},~\ref{fig:IVIFmetric},~\ref{fig:medical}, and~\ref{fig:13}. To further evaluate the robustness of the proposed method in challenging conditions, such as smoke occlusion and imaging noise, we conducted additional analysis experiments.

In smoke occlusion scenarios, as shown in Figure~\ref{fig:16}, most existing methods effectively preserve salient information from infrared images. However, ITFuse and TarDAL are severely compromised by smoke interference during fusion, which nearly obscures the tree canopy and leads to significant loss of trunk detail, as particularly evident in the red zoomed-in box. Moreover, as observed in both the red and green zoomed-in boxes, all methods except ours inappropriately fuse the background information from the visible image, resulting in perceptually unrealistic effects. In contrast, our method demonstrates superior performance in terms of color fidelity. Achieving an optimal equilibrium between preserving prominent targets and retaining contextual environmental details (such as weather conditions) in fused imagery continues to pose a substantial unresolved research challenge.

In noise scenarios, we follow the approach described in~\citep{DRMF} to introduce various types of noise into the input images to simulate common interference factors encountered in real-world imaging systems. Specifically, for the IVIF task, random stripe noise with variance $\sigma^2=5$ is added to the infrared images, while Gaussian noise with variance $\sigma^2=25$ is introduced into the visible images. For the MIF task, Gaussian-Poisson noise is applied to SPECT images, and random Gaussian noise with $\sigma^2=25$ is added to MRI images. As shown in Figure~\ref{fig:17}, most comparative methods (e.g., GTF, RFN-Nest, DDFM, and ITFuse) exhibit significant degradation in both salient target representation and background clarity. In particular, ITFuse and TarDAL demonstrate severe susceptibility to the added noise, resulting in either over-smoothed or substantially distorted canopy structures within the red bounding box. Furthermore, fusion outcomes from methods such as DIDFuse and CMTFuse fails to preserve fine-grained background textures while introducing noticeable artifacts in vegetated regions (green box), yielding perceptually unrealistic results. In contrast, our proposed method exhibits robust noise suppression capabilities while simultaneously maintaining thermal target integrity and scene background fidelity. The human silhouettes within the red box remain sharply delineated, and the vegetation textures in the green box are preserved with superior color consistency and structural coherence. This validates our method's effectiveness in achieving balanced preservation of salient targets and environmental context under noisy conditions.

As further illustrated in Figure~\ref{fig:18}, methods including IFCNN, DDFM, ALFM, and TUFusion demonstrate insufficient handling of Poisson-Gaussian noise in SPECT images, leading to speckle artifacts (red box) and reduced contrast of MRI structural information. While EMFusion and MATR show improved noise robustness, they incur loss of fine SPECT structures (corresponding to MRI hypo-intense regions) within the red box. Comparatively, our method achieves optimal performance in preserving both functional SPECT information (hotspot regions in red box) and MRI anatomical details (ventricular margins in green box), with statistically significant noise suppression.

\textbf{Computational Complexity Analysis.} In this section, we evaluate computational efficiency from both time complexity and space complexity using six key metrics. The temporal metrics include model load time, average inference time (Avg Inftime), and floating-point operations (FLOPs). Model load time refers to the duration required to load a pre-trained model from persistent storage (e.g., disk or cloud) into memory, which is crucial for evaluating deployment latency. Average inference time directly reflects the processing speed and runtime efficiency of a model, indicating how quickly the model can complete a given task. FLOPs, representing the total number of floating-point operations executed during the forward pass, serve as a standard measure of computational complexity. A higher FLOPs value typically implies a more computationally intensive model, which requires increased processing resources. 
The spatial metrics comprise the number of trainable parameters, model size, and peak GPU memory consumption (Peak Memory). The parameter count, encompassing all learnable weights and biases, provides insight into the model’s capacity and structural complexity. Model size denotes the storage space occupied by the entire model, which is a determining factor for its feasibility on resource-constrained or edge devices. Peak memory usage quantifies the maximum amount of GPU memory consumed throughout the model's execution, reflecting the upper bound of memory demand during inference. While models with larger parameter counts or memory footprints may offer enhanced representational power, they also incur higher computational and storage overheads.

For experimental validation, we randomly selected 100 images with a resolution of 640$\times$480 from the MSRS dataset to evaluate the proposed method in IVIF task, and another 100 images with a resolution of 256$\times$256 from the SPECT-MRI dataset to evaluate MIF task. 
All implementations are GPU-accelerated and executed on identical hardware.
The six evaluation metrics, summarized in Table~\ref{label:time-space}, are reported in milliseconds (ms), milliseconds (ms), gigaflops (G), thousand parameters (K), megabytes (MB), and megabytes (MB), respectively. 
To ensure fairness and minimize the influence of CPU performance, we utilized third-party open-source libraries for metric computation.

As shown in Table~\ref{table:time_memory}, the proposed method performs relatively well across multiple computational metrics. Specifically, its average inference time reaches 149.704 ms, which is significantly higher than that of lightweight models such as U2Fusion (24.747 ms) and TarDAL (14.792 ms), but still considerably lower than that of DDFM (500.353k ms), demonstrating a certain degree of computational efficiency. Its FLOPs are relatively high among all methods, indicating a higher computational complexity. This also indirectly suggests that our model is capable of extracting and integrating multimodal features from infrared and visible images more effectively. Furthermore, in the MIF task, our method shows a similar trend in computational efficiency and space consumption. Although it consumes slightly more resources compared to lightweight models, it significantly outperforms other methods in terms of fusion quality. This clearly reflects the effectiveness of the ``performance-for-computation” strategy and validates the practical utility and effectiveness of the proposed architecture in enhancing fusion quality.

\textbf{Experiment on $\alpha$ and $\beta$.}
The total loss $\mathcal{L}_{total}$ consists of three components: $\mathcal{L}_{str}$, $\mathcal{L}_{rec}$, and $\mathcal{L}_{fus}$. For $\mathcal{L}_{fus}$, the weights of the sub-loss terms, $\mathcal{L}_{ssim}$, $\mathcal{L}_{smooth}$, and $\mathcal{L}_{grad}$, are empirically set to 1~\citep{IAIFNet, UMF-CMGR}. As for $\mathcal{L}_{str}$ and $\mathcal{L}_{rec}$, we conduct a parameter sensitivity analysis on their associated hyperparameters $(\alpha, \beta)$, with detailed results provided in Table~\ref{table6}.

\begin{table*}[h]
\centering
\caption{Time and Space Complexity Comparison of IVIF Methods on the MSRS Dataset and MIF Methods on the SPECT-MRI Dataset.}
\label{table:time_memory}
%\resizebox{\textwidth}{!}{%
\begin{tabular}{c|c|ccc|ccc}
\toprule
& \multirow{2}{*}{\centering \textbf{Method}}& \multicolumn{3}{c|}{\textbf{Time complexity}} & \multicolumn{3}{c}{\textbf{Space complexity}} \\
& & \text{Load Time}\,\raisebox{-0.3ex}{\scriptsize(ms)}& \text{Avg Inftime}\,\raisebox{-0.3ex}{\scriptsize(ms)}
 & \text{FLOPs}\,\raisebox{-0.3ex}{\scriptsize(G)}
 & \text{Parameters}\,\raisebox{-0.3ex}{\scriptsize(K)}
& \text{Model Size}\,\raisebox{-0.3ex}{\scriptsize(MB)}
 &\text{Peak Memory}\,\raisebox{-0.3ex}{\scriptsize(MB)}
 \\
\midrule
\multirow{9}{*}{\centering IVIF}&GTF      & ——       & 0.639     & 0.052    & 0.016       & ——        & —— \\
&DIDFuse  & 6.189    & 19.408    & 114.86   & 260.935     & 0.998     & 1754.565 \\
&AUIF     & 7.405    & 41.187    & 0.065    & 11.631      & 0.045     & 172.381 \\
&RFN-Nest & 277.187  & 75.547    & 138.636  & 7524.249    & 28.703    & 2290.235 \\
&U2Fusion & 1.062    & 24.747    & 202.361  & 659.217     & 2.515     & 931.435 \\
&DeFusion & 14.582   & 45.883    & 80.138   & 7874.473    & 30.039    & 666.17 \\
&UMF      & 742.162  & 28.726    & 193.207  & 629.253     & 2.4       & 1079.354 \\
&TarDAL   & 0.486    & 14.792    & 91.138   & 296.577     & 1.133     & 11145.65 \\
&ITFuse   & 2.252    & 100.062   & 26.603   & 82.487      & 0.318     & 1173.559 \\
&CMTF     & 5.676    & 49.777    & 61.421   & 623.989     & 2.387     & 1144.485 \\
&DDFM     & 3499.827 & 500.353k   & 5.220k     & 552814.086  & 2108.818  & 4863.326 \\
&Ours     & 2.265    & 149.704   & 1.332k & 12021.649   & 45.859    & 3730.944 \\
\midrule
\multirow{5}{*}{\centering MIF} &IFCNN    & 587.050   & 1.330    & 8.544   & 74.179    & 0.320    & 88.040 \\
&EMFusion & 9.889  & 20.065    & 3.957  & 148.536     & 4.666  & 95.623 \\
&DDFM     & 3460.532 & 14.587K   & 1.114k     & 552814.086  & 2108.818  & 4811.36 \\
&MATR     & 1.145    & 27.450    & 1.948   & 13.422     & 0.052     & 86.921 \\
&ALMF     & 2.233 & 4.332   & 3.583     & 53.719  & 0.208  & 89.797 \\
&TUFusion & 9.357    & 109.036   & 13.428   & 19070.503      & 72.748     & 1361.924 \\
&Ours     & 2.205    & 3.815k   & 284.153 & 12021.649   & 45.859    & 964.659 \\
\bottomrule
\end{tabular}%
\label{label:time-space}
%}
\end{table*}

\begin{table}[t]
	\begin{center}
\caption{Quantitative comparison between manually crafted constant weights on two datasets in terms of $\alpha$ and $\beta$.}
\begin{tabular}{cc|cccccc}
\toprule
$\mathcal{\alpha}$ & $\mathcal{\beta}$ & MI & SF & AG & VIF & Qabf & SSIM \\
\midrule
\multicolumn{8}{c}{\textbf{MSRS}} \\
\midrule
0.1 & 0.1 & 3.564 & 10.523 & 3.153 & 0.786 & 0.775 & 0.992 \\
1.0 & 0.5 & 4.568 & 10.554 & 3.224 & 0.965 & 0.675 & 0.956 \\
0.5 & 1.0 & 5.760 & 10.576 & 3.236 & 1.014 & \textbf{0.805} & 0.923 \\
1.0 & 1.0 & \textbf{5.792} & \textbf{11.007} & \textbf{3.499} & \textbf{1.035} & 0.703 & \textbf{0.993} \\
\midrule
\multicolumn{8}{c}{\textbf{SPECT-MRI}} \\
\midrule
0.1 & 0.1 & 3.461 & 18.160 & 6.433 & 0.726 & \textbf{0.760} & 0.932 \\
1.0 & 0.5 & 3.275 & 18.081 & 6.558 & 0.765 & 0.635 & 0.995 \\
0.5 & 1.0 & 4.001 & 19.730 & \textbf{7.252} & \textbf{0.830} & 0.743 & 1.122 \\
1.0 & 1.0 & \textbf{4.118} & \textbf{22.919} & 7.199 & 0.822 & 0.756 & \textbf{1.225} \\
\bottomrule
\end{tabular}
\label{table6}
\end{center}
\end{table}

Table~\ref{table6} presents performance of the model under different weight combinations $(\alpha, \beta)$ across six evaluation metrics on the MSRS and SPECT-MRI datasets, aiming to analyze the sensitivity of the loss function to these weight parameters. The results indicate that metrics such as MI, SF, and AG are relatively sensitive to the weight settings, showing noticeable performance fluctuations as $\alpha$ and $\beta$ vary. In contrast, SSIM exhibits low sensitivity to weight changes, demonstrating high robustness. On both datasets, the combination $(\alpha = 1.0, \beta = 1.0)$ generally yields the best or near-best overall performance, suggesting that this setting achieves a good balance across multiple metrics. Additionally, the setting $(\alpha = 0.5, \beta = 1.0)$ performs well in perceptual-related metrics such as AG, Qabf and VIF, making it suitable for scenarios that emphasize visual quality. These sensitivity analysis provides quantitative guidance for the rational selection of weight parameters in loss function.

\section{Conclusion and Discussion}\label{sec:5}
\label{sec:typestyle}
In this paper, we propose a novel semantic structure-preserving approach for multi-modality image fusion. The whole fusion paradigm consists of two major modules. To be more specific, we design a structural feature extractor in our network to extract the salient structural feature representation. Moreover, we develop a structure-preserving fusion module for integrating multi-scale structural features to ensure consistent semantic structure across multiple modalities and prevent information loss during the fusion process. The mutual coordination of these two modules results in structure-aware consistency throughout image fusion, without sacrificing the visual quality and color fidelity of the fused image.
We conduct extensive experiments to validate that our method is capable of handling either IVIF or MIF tasks in various illumination conditions. The experimental results further demonstrate that our method has advantages over nine state-of-the-art methods.

\textbf{Limitations and Future Work}. Despite the promising results, there remain opportunities to further enhance SSPFusion's performance and applicability. Future work will focus on two key directions: (1) Optimizing memory usage during training and inference to alleviate the resource-intensive challenges of deploying it on micro hardware.
(2) Extending the proposed method to other computer vision tasks, such as object detection, to enhance its practical utility. Although we have only demonstrated the advantages of SSPFusion in semantic segmentation, we believe it is applicable to broader applications.
In the future, we plan to explore the potential of multimodal image fusion in other application areas, aiming to push the boundaries of image fusion research and broaden the real-world impact of SSPFusion.

\section*{Acknowledgments}
This work was supported by the National Natural Science Foundation of China (No.61976017, No.61601021, and No. 62132002), the Beijing Natural Science Foundation (No.4202056 and No.L172022) and the Fundamental Research Funds for the Central Universities (2022JBMC013, 501QYJC2024115010 and 502GWXM2024115007).

%% Loading bibliography style file
%\bibliographystyle{model1-num-names}
\bibliographystyle{cas-model2-names}

% Loading bibliography database
\bibliography{reference}

\end{document}